%% file: root.tex
\documentclass[letterpaper, 10 pt, conference]{ieeeconf}  

\IEEEoverridecommandlockouts                              

\usepackage[numbers]{natbib}
\usepackage{graphicx}
\usepackage{amsmath}
\usepackage{hyperref}
\usepackage{algpseudocode}
\usepackage{algorithm}
\usepackage{multirow}
\usepackage[normalem]{ulem}
\usepackage{booktabs}
\usepackage{lipsum}

\usepackage[utf8]{inputenc} 
\usepackage[T1]{fontenc}    
\usepackage{url}            
\usepackage{amsfonts}       
\usepackage{nicefrac}       
\usepackage{microtype}      
\usepackage{xcolor}         
\usepackage{amssymb}         
\usepackage{color}
\usepackage{blindtext}
\let\labelindent\relax
\usepackage{enumitem}
\usepackage{arydshln}
\usepackage{wrapfig}
\usepackage{bbm}
\usepackage{breqn}

\input{math_commands.tex}

\newcommand{\method}{{PolyTask}}
\newcommand{\learn}{\textit{learn}} 
\newcommand{\distill}{\textit{distill}}

\hypersetup{
    colorlinks=true,
    linkcolor=blue,
    filecolor=magenta,      
    citecolor=blue,
    urlcolor=blue,
}

\newcommand{\xxnote}[3]{}
\ifx\hidenotes\undefined
  \renewcommand{\xxnote}[3]{} 
\fi

\newcommand{\SH}[1]{{\xxnote{SH}{red}{#1}}}

\usepackage{tikz}
\newcommand{\cblock}[3]{
  \hspace{-1.5mm}
  \begin{tikzpicture}
    [
    node/.style={square, minimum size=10mm, thick, line width=0pt},
    ]
    \node[fill={rgb,255:red,#1;green,#2;blue,#3}] () [] {};
  \end{tikzpicture}%
}

\title{\method{}: Learning Unified Policies through \\ Behavior Distillation}

\author{Siddhant Haldar\thanks{Correspondence to: siddhanthaldar@nyu.edu} and Lerrel Pinto \vspace{0.1in}
\\ \vspace{0.1in} New York University
\\{\small \tt \href{https://poly-task.github.io/}{poly-task.github.io}}
\vspace{-0.2in}
}

\newboolean{include-notes}
\setboolean{include-notes}{true}
\newcommand{\SC}[1]{\ifthenelse{\boolean{include-notes}}%
{\textcolor{green}{\textbf{SC: #1}}}{}}
\newcommand{\BE}[1]{\ifthenelse{\boolean{include-notes}}%
{\textcolor{green}{\textbf{BE: #1}}}{}}

\begin{document}


\maketitle
\thispagestyle{empty}
\pagestyle{empty}

\begin{abstract}
\input{documents/abstract}
\end{abstract}

\section{Introduction}

\input{documents/introduction}

\section{\method{}}
\input{documents/approach}

\section{Experiments}
\input{documents/experiments}

\section{Related Work}
\input{documents/related_work}

\section{Conclusion and Limitations}
\input{documents/limitations}

\section{Acknowledgements}
\input{documents/acknowledgements}

\bibliographystyle{IEEEtran}
\small
\bibliography{references}


\clearpage
\appendix
\input{documents/appendix}

\end{document}

%% file: math_commands.tex
\usepackage{amsmath,amsfonts,bm}
\usepackage{caption}
\usepackage{subcaption}

\def\eqref#1{equation~\ref{#1}}

\def\1{\bm{1}}

\DeclareMathAlphabet{\mathsfit}{\encodingdefault}{\sfdefault}{m}{sl}
\SetMathAlphabet{\mathsfit}{bold}{\encodingdefault}{\sfdefault}{bx}{n}

\def\gT{{\mathcal{T}}}



%% file: documents/abstract.tex
Unified models capable of solving a wide variety of tasks have gained traction in vision and NLP due to their ability to share regularities and structures across tasks, which improves individual task performance and reduces computational footprint. However, the impact of such models remains limited in embodied learning problems, which present unique challenges due to interactivity, sample inefficiency, and sequential task presentation. In this work, we present PolyTask, a novel method for learning a single unified model that can solve various embodied tasks through a `learn then distill' mechanism. In the `learn' step, PolyTask leverages a few demonstrations for each task to train task-specific policies. Then, in the `distill' step, task-specific policies are distilled into a single policy using a new distillation method called \textit{Behavior Distillation}. Given a unified policy, individual task behavior can be extracted through conditioning variables. PolyTask is designed to be conceptually simple while being able to leverage well-established algorithms in RL to enable interactivity, a handful of expert demonstrations to allow for sample efficiency, and preventing interactive access to tasks during distillation to enable lifelong learning. Experiments across three simulated environment suites and a real-robot suite show that PolyTask outperforms prior state-of-the-art approaches in multi-task and lifelong learning settings by significant margins.

%% file: documents/introduction.tex

Current progress in large-scale machine learning has been driven by large unified models that can solve multiple tasks~\cite{reed2022generalist,driess2023palm,shridhar2023perceiver,brohan2022rt}. In contrast to task-specific models, unified models are hypothesized to benefit from sharing data, regularizing representations, and reducing overall parameter counts~\cite{lu2022unified,kolesnikov2022uvim,Casas_2021_CVPR,dong2019unified,girdhar2023imagebind}. Perhaps more importantly, having a single model streamlines the assimilation of skills by circumventing challenges associated with managing numerous skills during deployment. While we have built better frameworks to learn unified models in vision~\cite{lu2022unified,kolesnikov2022uvim,Casas_2021_CVPR,girdhar2023imagebind} and natural language processing~\cite{dong2019unified,lu2022unified,bao2020unilmv2}, their impact in embodied domains -- where agents must solve problems by interacting with physical environments -- has been limited. This is despite tremendous improvement in single-task policy learning in recent years~\cite{uchendu2022jump,jena2020augmenting,haldar2023watch,haldar2023teach,smith2022walk,ball2023efficient}.


Prior efforts into training unified policies fall under the umbrella of multi-task policy learning and can be broadly categorized into two paradigms -- offline imitation and online reinforcement learning (RL). Offline imitation learning uses large amounts of expert demonstrations to supervise the training of the unified policy. However, training such unified policies often requires several thousand demonstrations to learn simple control tasks~\cite{reed2022generalist}. Multi-task RL~\cite{sodhani2021multi,yu2020gradient,teh2017distral} approaches on the other hand do away with the need for demonstrations and instead use reward-driven learning to optimize the unified policy. This comes at the cost of large amounts of interactive experience, often exceeding the cumulative sum of experience to train individual task-specific policies~\cite{vithayathil2020survey}.

\begin{figure}[t!]
    \centering
    \includegraphics[width=\linewidth]{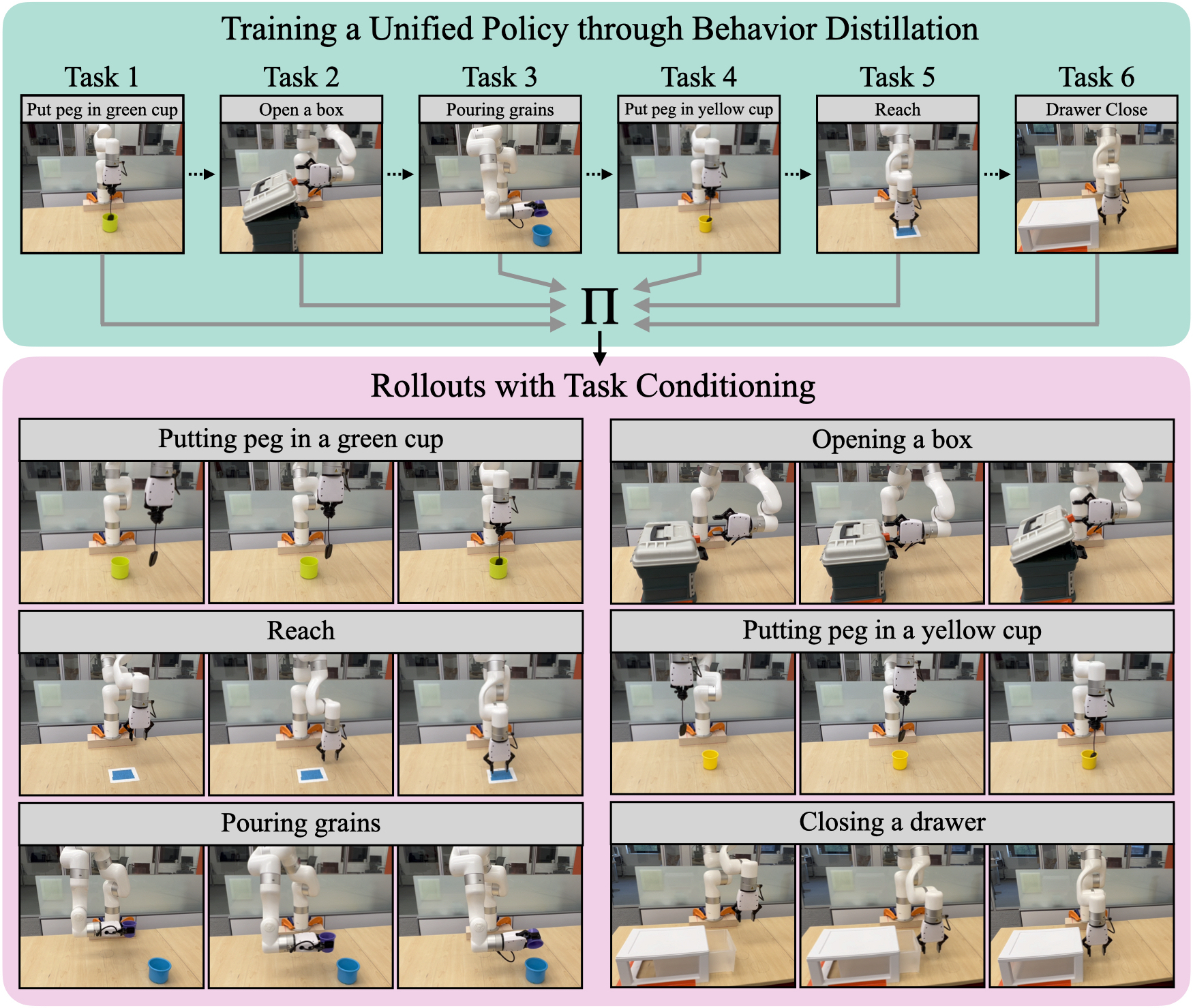}    
    \caption{\method{} is a technique to train a single unified policy $\Pi$ that can solve a range of different tasks. This is done by first learning a single-task policy for each task followed by distilling it into a unified policy. Distillation allows our unified policy to assimilate additional tasks in a lifelong-learning fashion without needing to increase the parameter count of the policy. Once trained, the unified policy can solve tasks by conditioning on task identifiers such as goal image, text description, or one-hot labels.}
    \label{fig:intro}
\end{figure}

A more nuanced challenge in training a unified policy with RL is the need to access the environment of all tasks simultaneously in parallel. While this is easy to do with stationary datasets~\cite{reed2022generalist,driess2023palm}, collecting interactive experiences in a multitude of environments at the same time is only feasible in simulation. Real-world embodied agents, such as robots can only be solving one task at a time, which brings about challenges in catastrophically forgetting prior tasks~\cite{robins1993catastrophic}. 


In this work, we present \method{}, a new framework to train unified policies that can solve a multitude of tasks. \method{} is intuitively simple, requires a minimal increase in parameter counts compared to a single-task policy, and is readily applicable to a wide variety of policy learning settings. At its core, \method{} is built on the principle of `learn then distill', a two-phase process. In the `learn' phase, a single-task policy is trained using demonstration-guided RL for every task. This training procedure combines the sample efficiency of learning from demonstrations with the flexibility of interactive RL. Next, we move to the `distill' phase. Here the single-task policies are distilled into a single unified policy using a new technique called \textit{Behavior Distillation}. Unlike prior work~\cite{rusu2015policy} that distill using Q-values, Behavior Distillation directly distills on policy outputs, which enables the distillation of continuous actions. With \method{}, we show for the first time that efficient demonstration-guided RL can be combined with multi-task distillation. This simple combination overcomes several prior challenges in multi-task policy learning by improving sample efficiency compared to multi-task RL and reducing the number of demonstrations needed compared to offline imitation.

\begin{figure}[t]
    \centering
    \includegraphics[width=0.8\linewidth]{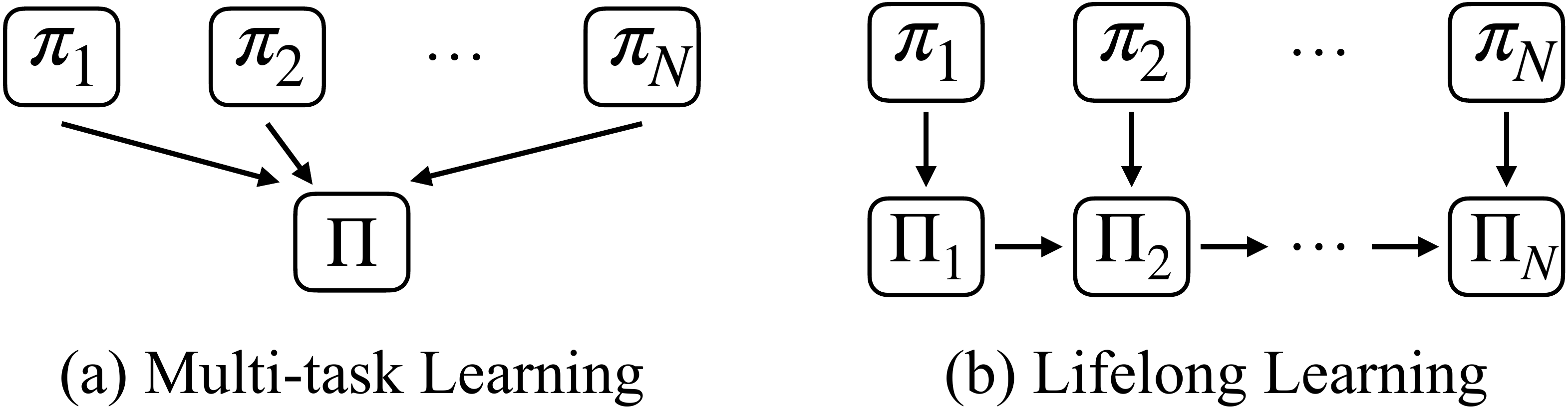}
    \caption{An illustration of the difference between training the unified policy on multi-task and lifelong learning settings.}
    \label{fig:schematic}
\end{figure}

Since Behavior Distillation only requires offline data, it makes \method{} readily applicable to lifelong learning settings, where tasks are presented sequentially. Concretely, when presented with a new task to solve, a new unified policy can be obtained by distilling both the old task experts and the new task policy into a single policy. This task assimilation is done without increasing parameter count while reducing catastrophic forgetting compared to prior work~\cite{julian2020efficient}. Such a lifelong distillation procedure lends itself well to real robotic problems where tasks are presented sequentially (see Fig. ~\ref{fig:intro}).



We evaluate \method{} on four environment suites -- MetaWorld, FrankaKitchen, DMControl, and xArm (real-robot) -- on multi-task and lifelong settings to learn unified policies. Through an extensive study of 32 tasks across 3 simulation suites and 6 tasks on a xArm robotic arm, we present the following key insights:
\begin{enumerate}
    \item \method{} outperforms prior state-of-the-art multi-task learning algorithms by an average of 1.3$\times$ on 32 tasks across 3 simulated environment suites (Sec.~\ref{subsec:mt_exps}).
    \item Behavior Distillation allows \method{} tackle catastrophic forgetting in lifelong learning without increasing parameter counts, yielding a 3.8$\times$ improvement over prior work (Sec.~\ref{subsec:ll_exps}).
    \item On our real-robot benchmark, we find that \method{} can be used for both multi-task policy learning and lifelong learning. In both settings, our single unified policy performs on par with the task-specific experts without needing additional data for training (Sec.~\ref{subsec:real_robot}).
    \item Through an ablation analysis, we demonstrate that \method{} can also work without environment rewards, is not sensitive to the size of the policy network, and works with various conditioning modalities including one-hot, language, and goal-images (Sec.~\ref{subsec:ablations}).
\end{enumerate}


%% file: documents/approach.tex
\label{sec:approach}

\subsection{Problem formulation and overview}
\label{subsec:problem_formulation}
A fundamental challenge in embodied learning problems is to develop a single unified model that can solve a variety of tasks. Consider a set of $N$ tasks $\gT = \{\gT_1, \gT_2, \cdots, \gT_N \}$, each with a unique identifiable key $\{c_1, c_2, \cdots, c_N\}$. In such a setting, solving any particular task $\gT_k$ corresponds to maximizing the cumulative reward $r_k$ associated with it. We denote the task-specific policy for $\gT_k$ as $\pi_k$, which can be obtained using standard RL algorithms or through offline imitation, which requires expert demonstrations for each task. 


\paragraph{Multi-task learning} The goal of multi-task policy learning is to learn a single parametric policy $\Pi(a_t| o_t; c_k)$ that can solve a set of tasks $\gT$. During this learning, we assume parallel access to all tasks $\gT$, which allows for training on samples drawn on-policy for all tasks. Our work focuses on improving multi-task learning by reducing the number of samples needed by the RL policy while achieving high performance compared to single-task experts.

\paragraph{Lifelong learning of tasks} Given a multi-task policy $\Pi$ capable of performing the tasks in $\gT$, lifelong learning aims to teach a new set of tasks  $\gT^{\prime} = \{\gT^{\prime}_{1}, \gT^{\prime}_{2}, \cdots, \gT^{\prime}_{L}\}$ to policy $\Pi$ such that its effective set of tasks becomes $\gT^{\prime\prime} = \gT \bigcup \gT^{\prime}$ (shown in Fig.~\ref{fig:schematic}b). This comes with challenges related to transfer learning~\cite{ying2018transfer} and catastrophic forgetting~\cite{robins1993catastrophic,rusu2016progressive,parisi2017lifelong}. Our work primarily focuses on addressing catastrophic forgetting without needing additional policy parameters. 

\paragraph{Overview of \method{}} \method{} operates in 2 phases - \learn{} and \distill{}. During the \learn{} phase, a task-specific policy is learned for each task using offline imitation from a few expert demonstrations followed by online finetuning using RL. During the \distill{} phase, all task-specific policies are combined into a unified policy using knowledge distillation on the RL replay buffers stored for each task. 

\subsection{Phase 1: Learning individual task-specific policies}
\label{subsec:rl}
In this phase, a task-specific policy $\pi_k$ is learned for each task $\gT_k$. 
First, a randomly initialized policy is trained on a small set of expert demonstrations using goal-conditioned behavior cloning (BC)~\cite{lynch2020learning,emmons2021rvs}. This BC-pretrained policy is finetuned through online interactions in the environment using a standard RL optimizer. In this work, we use DrQ-v2~\cite{yarats2021mastering}, a deterministic actor-critic based method that provides high performance in continuous control. Following prior work in sample-efficient RL~\cite{haldar2023watch, jena2020augmenting, hoshino2022opirl}, we combine the online RL objective with a BC loss (as shown in Eq.~\ref{eq:rl}).

\begin{equation}
\begin{split}    
    \pi = \operatorname*{argmax}_\pi \left[(1-\lambda))\mathbb{E}_{(s,g,a)\sim \mathcal{D}^{\beta}}[Q(s,g,a)]\right.\\ \left.- \alpha \lambda\mathbb{E}_{(s^{e},g^{e},a^{e})\sim \mathcal{D}^{e}} \|a^{e} - \pi(s^{e}, g^{e})\|^{2} \right]
\end{split}
\label{eq:rl}
\end{equation}

Here, $Q(s,g,a)$ represents the Q-value from the critic used in actor-critic policy optimization with $g$ being the goal for the sampled state-action pair $(s,a)$. $\pi$ is a goal-conditioned policy. $\alpha$ is a fixed weight and $\lambda$ controls the relative contribution of the two loss terms. $\mathcal{D}^{\beta}$ refers to the replay buffer for online rollouts and $\mathcal{D}^{e}$  refers to the expert demonstration set. 

\subsection{Phase 2: Behavior-Distillation of multiple policies into a unified policy}
\label{subsec:distil}
Knowledge distillation~\cite{gou2021knowledge} is a method for transferring knowledge from a teacher model $T$ to a student model $S$. In addition to widespread use in vision and natural language processing~\cite{qu2021recent, xu2022survey, gou2021knowledge}, knowledge distillation has also been successful in policy learning~\cite{rusu2015policy, lai2020dual, czarnecki2019distilling}. Inspired by these prior works, we use knowledge distillation for combining the task-specific policies $\{\pi_1, ..., \pi_k\}$ obtained from the first phase into a unified goal-conditioned multi-task policy $\Pi$. 

In order to distill the knowledge from the previous phase, we use the replay buffer data obtained during the online RL training for each task. We cannot directly behavior clone the replay buffer data as this data is exploratory in nature and hence suboptimal. To tackle this, \cite{rusu2015policy} propose distilling the Q-values for $(s, a)$ tuples in the task-specific replay buffers into the unified policy. This works well for discrete action control as the distilled Q-function can be converted to a policy through the argmax operation. 
However, for continuous control problems, such Q-value distillation is incompatible as the argmax operation requires additional optimization to produce executable policies~\cite{ryu2019caql}.

Instead, we propose behavior distillation, a new and simple technique to distill policies without needing access to Q values. We directly use the learned task-specific teacher policy to relabel the action corresponding to each replay buffer state and distill the action distribution of the relabeled actions into the unified policy. Our distillation objective has been shown in Eq.~\ref{eq:distillation}.

\begin{equation}
    \Pi = \operatorname*{argmin}_{\Pi}\mathbb{E}_{t \sim \gT}\mathbb{E}_{(s,g)\sim \mathcal{D}^{\beta}_{t}} \|\Pi(s,g) - \pi_{t}(s,g)\|^{2}
    \label{eq:distillation}
\end{equation}

Here, $\gT$ refers to the set of tasks and $\mathcal{D}^{\beta}_{t}$ refers to the replay buffer for the task-specific policy $\pi_{t}$ for all tasks $t\in\gT$. $\Pi$ is the unified multi-task policy learned through distillation. 

\subsection{Extending \method{} to lifelong learning with sequential task presentation}
\label{subsec:ll}
Lifelong learning refers to a scenario where a robot encounters various tasks in a sequential manner, while being exposed to only one task at a given moment. In this setting, the state distributions change over time, which often leads to policies catastrophically forgetting previous tasks~\cite{robins1993catastrophic}. As \method{} distills policy only using off-policy data, it naturally fits lifelong settings without much modification. Unlike prior work that continuously finetune on new tasks~\cite{julian2020efficient} or train separate networks for each new task~\cite{rusu2016progressive}, \method{} uses prior task data to prevent forgetting while using the same model architecture to prevent an explosion of parameters.  

For every task-specific expert policy $\pi_{i}$ learned using demonstration-guided RL, each $(s,g,a)$ sample in the replay buffer $\mathcal{D}^{\beta}_{i}$ is replaced by $(s,g,\pi_{i}(s,g))$. This is done to have the unified policy model expert actions, not sub-optimal training actions. Given policies $\pi_{1:N}$ corresponding to tasks $\gT_{1:N}$, we first distill the knowledge of $\pi_{1:N}$ into a unified policy $\Pi_N$ using the relabeled task-specific replay buffers $\mathcal{D}^{\beta}_{1:N}$ (using Eq.~\ref{eq:distillation}). Next, in order to teach a new task $\gT_{N+1}$, we relabel the task-specific replay buffer $\mathcal{D}^{\beta}_{N+1}$ using the task expert $\pi_{N+1}$. Finally, a unified policy $\Pi_{N+1}$ for tasks $\gT_{1:N+1}$ is obtained through the same distillation procedure applied on replay buffers $\mathcal{D}^{\beta}_{1:N+1}$.





%% file: documents/experiments.tex
\label{sec:experiments}
Our experiments are designed to answer the following questions: $(a)$ How well does \method{} work in multi-task distillation? $(b)$ Can \method{} deal with the sequential presentation of tasks? $(c)$ Does \method{} scale to real-world robots? $(d)$ What design decisions in \method{} affect performance?

\subsection{Experimental setup}
\label{subsec:experimental_setup}

\begin{figure}
    \centering
    \includegraphics[width=\linewidth]{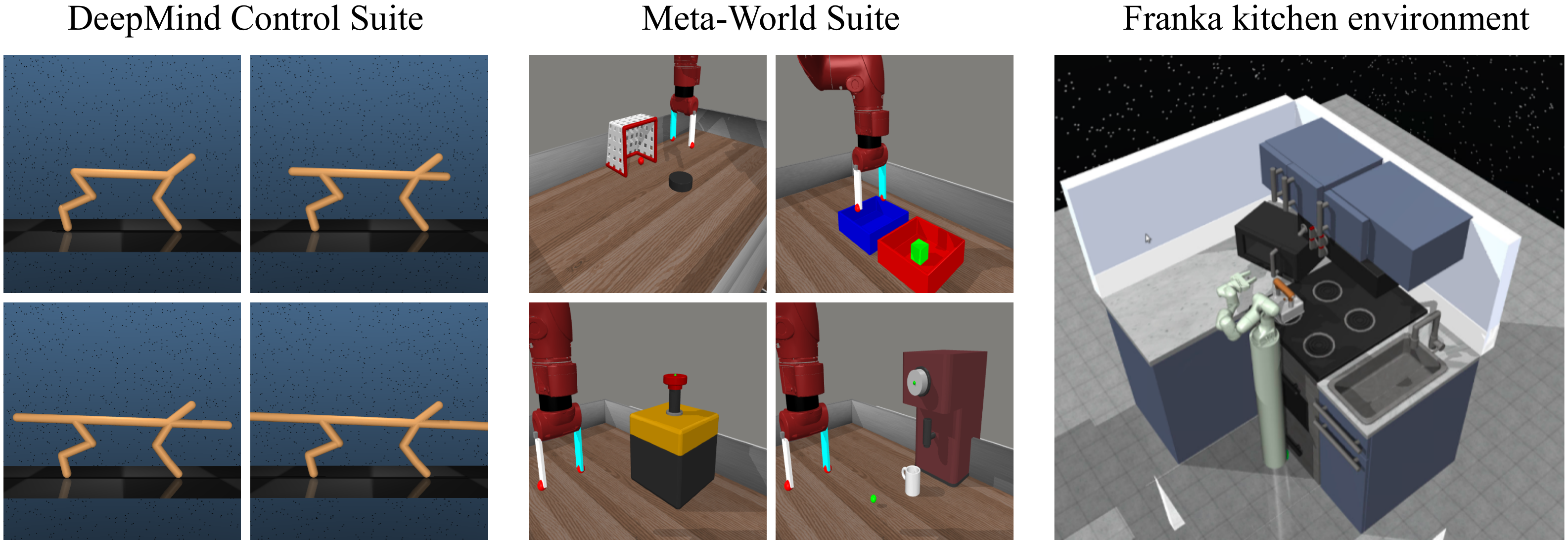}
    \caption{\method{} is evaluated across 3 simulated benchmarks - the DeepMind Control suite, the Meta-World benchmark, and the Franka kitchen environment.}
    \label{fig:sim_envs}
\end{figure}

\textbf{Simulated tasks:} We run experiments on 3 simulated environment suites across a total of 32 tasks.

\begin{enumerate}[leftmargin=*,align=left]
    \item \textbf{DM Control (DMC)}: We learn state-based policies spanning 10 tasks on the cheetah run environment in a multi-task variant of DM Control suite~\cite{tassa2018deepmind,todorov2012mujoco,Sodhani2021MTEnv}. Each task is a variation of the torso length of the cheetah. We train expert policies using DrQ-v2~\cite{yarats2021mastering} and collect 10 demonstrations for each task using this policy. A sinusoidal positional embedding~\cite{vaswani2017attention} corresponding to each task label is used as the goal embedding for these experiments.
    \item \textbf{Meta-World}: We learn image-based policies spanning 16 tasks from the Meta-World suite~\cite{yu2019meta}. For each task, we collect a single hard-coded expert demonstration from their open-source implementation~\cite{yu2019meta}. The last frame of the demonstration is used as the goal for each task.
    \item \textbf{Franka kitchen}: We learn image-based policies spanning 6 tasks from the Franka kitchen environment~\cite{gupta2019relay}. We use the relay policy learning dataset~\cite{gupta2019relay} comprising 566 demonstrations. Since each trajectory in the dataset performs multiple tasks, we segment each trajectory into task-specific snippets and use 100 demonstrations for each task. It must be noted that since the relay policy learning dataset~\cite{gupta2019relay} consists of play data, the segmented task-specific snippets are suboptimal which makes the learning problem harder. For each online rollout, we randomly select one of the task demonstrations and use the last frame as the goal image.
\end{enumerate}


\textbf{Robot tasks}: Our real-world setup comprises six tasks as shown in Fig.~\ref{fig:intro}. We use a Ufactory xArm 7 robot with a xArm Gripper as the robot platform for our real-world experiments. The observations are RGB images from a fixed camera. In this configuration, we gather up to two demonstrations per task and proceed to train our task-specific expert policies through demonstration-guided RL, employing rewards based on optimal transport (OT) based trajectory matching~\cite{haldar2023watch, haldar2023teach, cohen2022imitation}. We limit the online training to a fixed period of 1 hour. 

\textbf{Baseline methods}: We compare \method{} to a variety of baselines in multi-task policy learning and lifelong learning. A brief discussion of them is as follows:

\begin{enumerate}[leftmargin=*,align=left]
    \item \textbf{Goal conditioned BC (GCBC)}~\cite{lynch2020learning,emmons2021rvs}: A supervised learning framework for learning a goal-conditioned multi-task policy $\pi(\cdot| o, g)$ given a dataset of (observation, action, goal) tuples $(o, a, g)$.
    \item \textbf{Multi-task RL (MTRL)}~\cite{vithayathil2020survey,teh2017distral,sodhani2021multi, kalashnikov2021mt}: A framework for learning multi-task policies where the agent is simultaneously trained on all tasks using reinforcement learning.
    \item \textbf{Distral}~\cite{teh2017distral}: A MTRL algorithm that jointly trains separate task-specific policies and a distilled policy that captures common behavior across tasks. 
    \item \textbf{MTRL-PCGrad}~\cite{yu2020gradient}: A variant of MTRL with {projecting-conflicting-gradients (PCGrad)} style gradient optimization to mitigate task interference during gradient updates.
    \item \textbf{MTRL-Demo}: A demo-guided variant of MTRL where we adapt the strategy proposed in Sec.~\ref{subsec:rl} to a multi-task setting. A multi-task BC policy is first trained on all task demonstrations and this policy is used for initialization and regularization during the MTRL training.
    \item \textbf{Fine-tuning}~\cite{julian2020efficient}: A framework where a single policy is initialized and finetuned on each new task. The parameter count of the policy remains constant throughout training. 
    \item \textbf{Progressive Nets}~\cite{rusu2016progressive}: A framework that deals with catastrophic forgetting by adding a new model for each task with lateral connections to all previous models to allow for forward transfer. The parameter count of the policy increases with each new task. 
\end{enumerate}

GCBC, MTRL, Distral, MTRL-PCGrad, and MTRL-Demo serve as our multi-task skill learning baselines, and finetuning and progressive nets serve as our lifelong learning baselines.

\textbf{Evaluation metrics}: For each task across our environment suites, we measure performance by running 10 episodic seeds. Given this performance score on a task, we can then measure the effective number of tasks completed. The effective number of tasks executed by a policy is calculated as the cumulative success rate across all tasks in the entire task set. In the case of the Meta-World suite and Franka kitchen environment, where task completion is well-defined, we directly compute the metric using task success rate. However, in the cheetah run task of DM Control, the only available measure is the episode reward, with a maximum value of 1000. Thus, we calculate the average episode reward across 10 episodes, divide it by 1000, and consider that as the task success rate for all tasks.

\begin{table}[t]
\centering
\caption{Evaluation of multi-task distillation on 10 state-based tasks in DeepMind Control, 16 pixel-based tasks in the Meta-World benchmark, and 6 pixel-based tasks in the Franka kitchen environment. We notice that \method{} performs a higher effective number of tasks as compared to prior work.}
\label{table:mt_results}
\begin{tabular}{lccc}
\hline
\textbf{Method}                   & \textbf{DMC} & \textbf{Meta-World} & \textbf{Franka Kitchen} \\ \hline
Task-specific expert              &            7.31           &     15.6            &           4.8           \\ \hdashline
GCBC~\cite{lynch2020learning,emmons2021rvs}                   &            1.76           &     6.3             &           1.7           \\
MTRL~\cite{vithayathil2020survey,teh2017distral,sodhani2021multi, kalashnikov2021mt}  &            7.81           &     0.0             &           0.0           \\
Distral~\cite{teh2017distral}    &            0.08           &     0.9             &           N/A           \\
MTRL-PCGrad~\cite{yu2020gradient}                     &            8.09           &     1.0             &           N/A           \\
MTRL-Demo                         &      \textbf{8.79}        &     12.0            &           2.6           \\ \hdashline
\method{}                         &            7.12           &  \textbf{14.6}      &      \textbf{4.5}       \\
\method{}-PCGrad                  &            7.43           &     14.5            &           4.3           \\ \hline       
\end{tabular}
\vspace{5pt}
\end{table}

\begin{figure*}[t]
    \centering
    \includegraphics[width=1.0\linewidth]{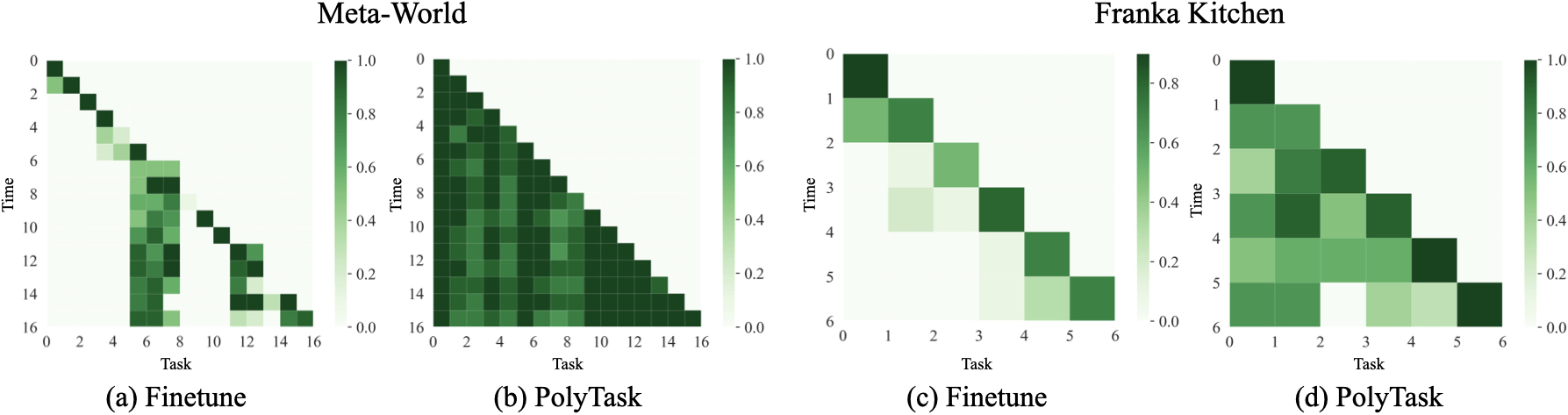}
    \caption{A comparison between the performance of Fine-tuning~\cite{julian2020efficient,xie2022lifelong} and \method{} on the Meta-World benchmark [(a), (b)] and the Franka kitchen environment [(c), (d)] in a lifelong learning setting. We observe that \method{} exhibits a significantly better ability to tackle catastrophic forgetting.}
    \label{fig:finetune_lifelong}
\end{figure*}

\begin{figure}
    \centering
    \includegraphics[width=\linewidth]{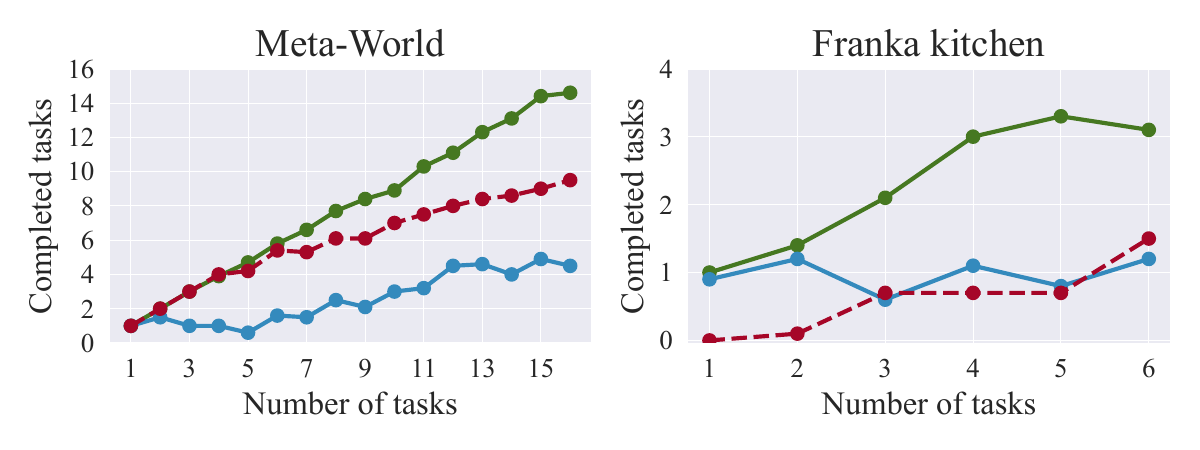}
    \hspace{6mm}\cblock{166}{6}{40}\hspace{1mm}Progressive Networks\hspace{1.5mm}
    \cblock{52}{138}{189}\hspace{1mm}Finetuning\hspace{1.5mm}
    \cblock{70}{120}{33}\hspace{1mm}\method{} (Ours)\hspace{1.5mm}
    \caption{Pixel-based evaluation for lifelong learning on 16 tasks in Meta-World, and 6 tasks in Franka kitchen.}
    \label{fig:ll_results}
\end{figure}

\subsection{How well does \method{} work in multi-task distillation?}
\label{subsec:mt_exps}
Table~\ref{table:mt_results} shows our results for multi-task policy learning across 3 simulated environment suites. We provide the results for two variants of \method{} - a vanilla version using Adam optimizer~\cite{kingma2014adam} and \method{} with PCGrad~\cite{yu2020gradient} based gradient optimization which is aimed at aiding multitask learning. On the simpler cheetah run task in DeepMind Control, we observe that MTRL-based baselines outperform \method{}. This can be attributed to the fact that with such low environment variations (only the torso length of the cheetah in this case), sharing information between tasks enables more robust learning. However, since we use knowledge distillation, \method{} shows a performance at par with the task-specific experts. 

On the harder tasks in Meta-World and Franka Kitchen, we see a bigger gap in performance with \method{} performing $2.3\times$ better than GCBC and $1.3\times$ better than the strongest MTRL-Demo baseline. We also present results on \method{}-PCGrad and observe that the ``gradient surgery" proposed by PCGrad~\cite{yu2020gradient} achieves slightly better performance on DM Control while slightly under-performing on Meta-World and Franka-Kitchen. Due to PCGrad requiring a separate loss computation for each task, we ran into time constraints while running MTRL-PCGrad on the Franka kitchen environment and so we could not present this result. Further, the increase in effective performance from MTRL to MTRL-Demo highlights the importance of utilizing demonstrations while the increase in performance from MTRL-Demo to PolyTask shows the importance of distillation.

\subsection{Can \method{} deal with the sequential presentation of tasks (lifelong learning)?}
\label{subsec:ll_exps}
The offline knowledge distillation phase naturally makes \method{} suitable for lifelong learning. Fig.~\ref{fig:finetune_lifelong} and Fig.~\ref{fig:ll_results} demonstrate that \method{} exhibits significant lifelong learning capabilities as compared to fine-tuning a policy for the most recent task (a setting explored in ~\cite{julian2020efficient,xie2022lifelong}). On both Meta-World and Franka Kitchen, we observe that \method{} significantly outperforms finetuning with a sequential presentation of tasks. Progressive networks~\cite{rusu2016progressive} learn a new model for each new task, making them suitable for superior performance when handling a larger number of tasks. However, the performance comes at the cost of a significant increase in parameter count. Despite having a constant parameter count and the same overall environment interaction budget, we observe that \method{} outperforms progressive networks by a significant margin on both environment suites. The low performance of progressive networks can be accounted for by the inability of the RL method to perform the tasks.  For both finetuning and progressive networks, we add a behavior regularization term as proposed in Sec.~\ref{subsec:rl} in order to enhance sample efficiency.


\begin{table*}[t!]
\centering
\caption{\method{} is evaluated on a set of 6 robotic manipulations tasks. We notice the \method{} demonstrates remarkable multi-task and lifelong learning capabilities while achieving performance levels comparable to those of task experts.}
\label{table:real_results}
\begin{tabular}{lcccccc|c}
\hline
\textbf{Method}    & \textbf{\begin{tabular}[c]{@{}c@{}}Insert Peg in \\ Green Cup\end{tabular}} & \textbf{Open Box} & \textbf{Pour} & \textbf{\begin{tabular}[c]{@{}c@{}}Insert Peg in\\ Yellow Cup\end{tabular}} & \textbf{Reach} & \textbf{Drawer Close} & \textbf{\begin{tabular}[c]{@{}c@{}}Cumulative\\ success rate\end{tabular}}\\ \hline
Single task BC     &   4/10   &   0/10    &   3/10   &   5/10   &   3/10    &   5/10   &   2/6   \\
Single task expert &   5/10   &   10/10   &   6/10   &   9/10   &   10/10   &   10/10  &   5/6   \\
Multitask          &   7/10   &   10/10   &   6/10   &   9/10   &   10/10   &   10/10  &   5.2/6 \\
Lifelong           &   7/10   &   7/10    &   6/10   &   10/10  &   10/10   &   10/10  &   5/6 \\ \hline  
\end{tabular}
\end{table*}

\subsection{Does \method{} scale to real-world robots?}
\label{subsec:real_robot}

We devise a set of 6 manipulation tasks on a xArm robot to show the multi-task and lifelong learning capabilities of \method{} in the real world. For each task $\gT_k$, a task-specific expert $\pi_k$ is first trained using the scheme described in Sec.~\ref{subsec:rl}. For the multi-task experiments, we combine the expert policies into a unified policy using behavior distillation (as described in Sec.~\ref{subsec:distil}). For the lifelong experiments, we follow the scheme described in Sec.~\ref{subsec:ll} to sequentially combine the single-task expert policies. Table~\ref{table:real_results} shows the results on the real robot. We observe that the multi-task policy learned by ~\method{} is able to effectively perform 5.2 out of the 6 tasks and outperforms the single-task experts which effectively succeed on 5 out of 6 tasks. Further, in the task of inserting a peg in a green cup, we observe the single-task expert hovers over the cup and fails at placing the peg inside is some episodes. However, jointly training the unified policy seems to alleviate this issue (indicated by the higher success rate of multi-task and lifelong \method{}), hinting towards the benefits of parameter sharing and regularization when training a single unified model as opposed to multiple experts. In the lifelong learning setting, we observe that after training on the sequence of 6 tasks, \method{} is able to effectively perform 5 out of the 6 tasks, showing a significant ability to tackle catastrophic forgetting.

\subsection{What design decisions in \method{} affect performance?}
\label{subsec:ablations}
In this section, we ablate 3 design choices that have been made in \method{}.


\paragraph{Reward type} In our experiments on multi-task (Section~\ref{subsec:mt_exps}) and lifelong learning ( Section~\ref{subsec:ll_exps}), we initially assumed the availability of environment rewards for training task-specific experts. However, we found that this assumption can be relaxed, and even without environment rewards, we successfully applied inverse RL (IRL) with optimal transport (OT) based trajectory matching rewards, following prior work~\cite{haldar2023watch, haldar2023teach, cohen2022imitation}, to learn task-specific policies. These task experts are used to obtain a unified multi-task policy using \method{} for 10 tasks from the Meta-World benchmark, achieving a cumulative success rate of 8.2 across all tasks, matching individual task expert performance. Importantly, our real-world results in Table~\ref{table:real_results} also relied on OT-based rewards, confirming their effectiveness in the lifelong learning setting.

\paragraph{Network size} We analyze the impact of the distilled policy's size on \method{}. We evaluate the performance of \method{} in a lifelong learning setting using same task-specific experts as described in Sec.\ref{subsec:ll_exps} and with policies having parameter counts 0.1M, 0.5M, 1M, 4M(used throughout the paper), 10M, and 20M. Notably, we do not observe any significant performance differences attributed to the variation in policy size. This demonstrates the robustness of \method{} to smaller policy architectures.

\paragraph{Modality of goal conditioning} In this work, the modality used for representing goals has been images for the image-based tasks (Meta-World and Franka Kitchen) and a sinusoidal positional embedding~\cite{vaswani2017attention} for the state-based task (DeepMind Control). In this experiment, we evaluate the effect of different goal modalities on the multi-task performance of \method{}. We conduct our experiments on the 16 tasks in the Meta-World suite. We use the same task-specific experts with image-based goals as the expert policies. Only the goal modality of the distilled policy is changed. Table~\ref{table:goal_modality} provides the results of such multi-task learning with \method{} using one-hot goals and language-based goals as task identifiers. For the language-based goals, we encode the task labels for each task using a 6-layer version of MiniLM~\cite{wang2020minilm} provided in SentenceTransformers~\cite{reimers-2019-sentence-bert}. We observe that \method{} shows near-perfect performance with the simpler one-hot goal representation and the language-based goal specification outperforms the image-based goal specification. Thus, we conclude that \method{} can be used with various goal modalities, even when the goal modality for the distilled policy is different than the task-specific experts.

\begin{table}[H]
    \centering
    \caption{Results for \method{} with different goal modalities on 16 Meta-World tasks for multi-task distillation.}
    \label{table:goal_modality}
    \begin{tabular}{lccc}
        \hline
        \textbf{Goal Modality} & \textbf{Image} & \textbf{One-hot} & \textbf{Language} \\ \hline
        Meta-World  &  14.6  &  15.9  &  15.0  \\ \hline
    \end{tabular}
\end{table}

%% file: documents/related_work.tex
\subsection{Unified models}
Unified models have seen widespread success in computer vision~\cite{lu2022unified,kolesnikov2022uvim,Casas_2021_CVPR,girdhar2023imagebind}, natural language processing~\cite{dong2019unified,lu2022unified,bao2020unilmv2}, and decision-making~\cite{reed2022generalist,kalashnikov2021mt,shridhar2023perceiver,driess2023palm,wu2023mtm,brohan2022rt}. In this work, we focus on the application of unified models in decision-making. The unification in unified models can come in the form of a single model for multiple tasks~\cite{caruana1997multitask,kalashnikov2021mt,shridhar2023perceiver}, capable of handling multiple modalities~\cite{driess2023palm,girdhar2023imagebind,lu2022unified}, or being learned from diverse domains~\cite{reed2022generalist,majumdar2023we}. Though different in terms of application, all forms of unification hold the promise of enhancing decision quality, efficiency, and adaptability. In this work, we focus on multitask unification~\cite{shridhar2023perceiver,kalashnikov2021mt,reed2022generalist}. Inspired by recent advances in sample-efficient RL~\cite{uchendu2022jump,ball2023efficient,haldar2023watch,smith2022walk}, we propose a method for learning task-specific experts using online RL (Sec.~\ref{subsec:rl}) and unifying these experts using behavior distillation (Sec.~\ref{subsec:distil}).


\subsection{Multi-task RL} Multi-task RL (MTRL)~\cite{sodhani2021multi,teh2017distral,vithayathil2020survey,kalashnikov2021mt,parisotto2015actor,xu2020knowledge} is a branch of multi-task learning~\cite{caruana1997multitask} where a RL agent simultaneously performs multiple tasks to learn a multi-task policy for the set of tasks. MTRL is challenging due to interference between gradients of different tasks during gradient optimization and there has been some prior work aimed at mitigating this interference~\cite{yu2020gradient,chen2018gradnorm}. There have also been attempts at stabilizing MTRL through knowledge distillation~\cite{teh2017distral} and learning contextual representations~\cite{sodhani2021multi}. In Sec.~\ref{subsec:mt_exps}, we compare \method{} with a baseline MTRL-Demo which is inspired by recent advances in demonstration-guided RL~\cite{jena2020augmenting,uchendu2022jump,haldar2023watch,haldar2023teach,ball2023efficient} and provides a promising performance boost to prior MTRL approaches. Further, since the agent must simultaneously interact with multiple environments, MTRL is not feasible on a physical robot that can only access a single environment at any point in time.

\subsection{Lifelong learning} Lifelong learning~\cite{thrun1998lifelong,parisi2019continual} refers to the ability to continuously acquire and refine policies or decision-making strategies over time. Some of the challenges in lifelong learning are transfer learning~\cite{ying2018transfer}, catastrophic forgetting~\cite{robins1993catastrophic,rusu2016progressive,parisi2017lifelong}, incremental learning~\cite{rebuffi2017icarl} and the stability–plasticity dilemma~\cite{power2017neural,barnett2002and}. There have been some recent works tackling transfer learning by retaining prior experiences~\cite{xie2022lifelong}, learning a new model~\cite{rusu2016progressive} or a separate task head~\cite{li2017learning} for each task to avoid catastrophic forgetting, using inverse RL for lifelong learning~\cite{mendez2018lifelong} and using hyper-networks in conjunction with neural ODEs~\cite{chen2018neural} for learning robot policies~\cite{auddy2023continual}. Contrary to progressive networks~\cite{rusu2016progressive} that trains a separate model for each task to avoid catastrophic forgetting, \method{} uses prior task data while using the same model to avoid increasing the parameter count.

%% file: documents/limitations.tex
In this work, we present \method{}, a conceptually simple algorithm that unifies several task experts into a single multi-task policy through behavior distillation. We show the efficacy of our approach on a variety of simulated and robot domains. However, we recognize a few limitations in this work: (a) Since we store the replay buffer for each task, this approach might lead to storage concerns in the case of a large number of tasks. Extending \method{} to either avoid storing prior data or only storing a small number of data points using techniques such as reservoir sampling~\cite{vitter1985random} would be an interesting problem to tackle. (b) Though we use expert demonstrations to accelerate single-task learning, our framework currently does not allow forward transfer~\cite{parisi2017lifelong}. It would be interesting to see if enabling forward transfer in the \method{} framework can further improve performance.


%% file: documents/acknowledgements.tex
We thank Mahi Shafiullah for valuable feedback and discussions. This work was supported by grants from Google, Honda, Meta, Amazon, and ONR awards N00014-21-1-2758 and N00014-22-1-2773.

%% file: documents/appendix.tex
\section{Background}
\label{appendix:problem_setup}

\paragraph{Reinforcement Learning (RL)}We study RL as a discounted infinite-horizon Markov Decision Process (MDP). For pixel observations, the agent's state is approximated as a stack of consecutive RGB frames. The MDP is of the form $(\mathcal{O}, \mathcal{A}, P, R, \gamma, d_{0})$ where $\mathcal{O}$ is the observation space, $\mathcal{A}$ is the action space, $P: \mathcal{O}\times\mathcal{A}\rightarrow\Delta(\mathcal{O})$ is the transition function that defines the probability distribution over the next state given the current state and action, $R:\mathcal{O}\times\mathcal{A}\rightarrow\mathbb{R}$ is the reward function, $\gamma$ is the discount factor and $d_{0}$ is the initial state distribution. The goal is to find a policy $\pi:\mathcal{O}\rightarrow\Delta(\mathcal{A})$ that maximizes the expected discounted sum of rewards $\mathbb{E}_{\pi}[\Sigma_{t=0}^{\infty}\gamma^{t}R(\boldsymbol{o}_{t},\boldsymbol{a}_{t})]$, where $\boldsymbol{o}_{0}\sim d_{0}$, $a_{t}\sim \pi(\boldsymbol{o}_{t})$ and $\boldsymbol{o}_{t+1}\sim P(.|\boldsymbol{o}_{t},\boldsymbol{a}_{t})$.

\paragraph{Imitation Learning (IL)} The goal of imitation learning is to learn a behavior policy $\pi^b$ given access to either the expert policy $\pi^e$ or trajectories derived from the expert policy $\mathcal{T}^e$. While there is a multitude of settings with differing levels of access to the expert, this work operates in the setting where the agent only has access to observation-based trajectories, i.e. $\mathcal{T}^e \equiv \{(o_t, a_t)_{t=0}^{T}\}_{n=0}^N$. Here $N$ and $T$ denotes the number of trajectory rollouts and episode timesteps respectively. We choose this specific setting since obtaining observations and actions from expert or near-expert demonstrators is feasible in real-world settings and falls in line with recent work in this area~\cite{haldar2023watch,jena2020augmenting,uchendu2022jump}.

\paragraph{Optimal Transport for Imitation Learning~(OT)} In Sec.~\ref{subsec:real_robot} and Sec.~\ref{subsec:ablations}, we used optimal transport (OT) based inverse reinforcement learning (IRL) to learn robot policies in the real world. This is in line with an array of OT-based approaches that have recently been proposed for policy learning~\cite{haldar2023watch,haldar2023teach,cohen2022imitation}. Intuitively, the closeness between expert trajectories $\mathcal{T}^e$ and behavior trajectories $\mathcal{T}^b$ can be computed by measuring the optimal transport of probability mass from $\mathcal{T}^b \rightarrow \mathcal{T}^e$. During policy learning, the policy $\pi_{\phi}$ encompasses a feature preprocessor $f_{\phi}$ which transforms observations into informative state representations. Some examples of a preprocessor function $f_{\phi}$ are an identity function, a mean-variance scaling function and a parametric neural network. In this work, we use a parametric neural network as $f_{\phi}$. Given a cost function $c:\mathcal{O}\times\mathcal{O}\rightarrow\mathbb{R}$ defined in the preprocessor's output space and an OT objective $g$, the optimal alignment between an expert trajectory $\textbf{o}^{e}$ and a behavior trajectory $\textbf{o}^{b}$  can be computed as 

\begin{equation}
    \mu^{*} \in \underset{\mu\in\mathcal{M}}{\text{arg~min}}~g(\mu, f_{\phi}(\textbf{o}^{b}), f_{\phi}(\textbf{o}^{e}), c)
    \label{appendix:eq:alignment}
\end{equation}

where $\mathcal{M}=\{\mu\in\mathbb{R}^{T\times T}:\mu\boldsymbol{1}=\mu^{T}\boldsymbol{1}=\frac{1}{T}\boldsymbol{1}\}$ is the set of coupling matrices and the cost $c$ can be the Euclidean or Cosine distance. In this work, inspired by \cite{cohen2022imitation}, we use the entropic Wasserstein distance with cosine cost as our OT metric, which is given by the equation

\begin{equation}
\begin{aligned}
    g(\mu, f_{\phi}(\textbf{o}^{b}), f_{\phi}(\textbf{o}^{e}), c) &=  \mathcal{W}^{2}(f_{\phi}(\textbf{o}^{b}), f_{\phi}(\textbf{o}^{e}))\\
    &= \sum_{t,t'=1}^{T}C_{t,t^{'}} \mu_{t,t'}
\end{aligned}
\label{appendix:eq:wasserstein}
\end{equation}

where the cost matrix $C_{t,t^{'}} = c(f_{\phi}(\textbf{o}^{b}), f_{\phi}(\textbf{o}^{e}))$. Using Eq.~\ref{appendix:eq:wasserstein} and the optimal alignment $\mu^{*}$ obtained by optimizing Eq.~\ref{appendix:eq:alignment}, a reward signal can be computed for each observation using the equation
\begin{equation}
    \label{appendix:eq:ot_reward}
    r^{OT}(o^{\boldsymbol{b}}_{t}) = - \sum_{t'=1}^{T} C_{t,t^{'}} \mu^{*}_{t,t^{'}}
\end{equation}

Intuitively, maximizing this reward encourages the imitating agent to produce trajectories that closely match demonstrated trajectories. Since solving Eq.~\ref{appendix:eq:alignment} is computationally expensive, approximate solutions such as the Sinkhorn algorithm are used instead.


\begin{algorithm}
\caption{Demonstration-guided RL}\label{alg:demoRL}
\begin{algorithmic}
\State $\textbf{Require:}$\\
Expert Demonstrations $\mathcal{D}^e \equiv \{(\textbf{o}_t, \textbf{a}_t, \textbf{g}_t)_{t=0}^{T}\}_{n=0}^N$\\
Pretrained policy $\pi^{BC}$\\
Replay buffer $\mathcal{D}$, Training steps $T$, Episode Length $L$\\
Task environment $env$\\
Parametric networks for RL backbone (e.g., the encoder, policy, and critic function for DrQ-v2)\\
\\
\State $\textbf{Algorithm:}$
\State $\pi \gets \pi^{BC}$ \Comment{Initialize with pretrained policy}
\For{each timestep $t$ = $1...T$}
    \If{done}
        \State $\text{Add episode to}~\mathcal{D}$
        \State $\textbf{o}_{t} = env.reset(),~\text{done} = \text{False},~\text{episode} = [~]$ 
    \EndIf
    \State $\textbf{a}_{t} \sim \pi(\textbf{o}_{t}, \textbf{g}_{t})$
    \State $\textbf{o}_{t+1}, r_t, ~\text{done} = env.step(\textbf{a}_{t})$
    \State $\text{episode.append}([\textbf{o}_{t}, \textbf{a}_{t},\textbf{o}_{t+1}, r_{t}, \textbf{g}_{t}])$
    \State $\text{Update backbone-specific networks and}$
    \State $\text{reward-specific networks (for inverse RL) using}~\mathcal{D}$
\EndFor
\end{algorithmic}
\end{algorithm}

\begin{algorithm}
\caption{Behavior distillation for multi-task policy learning}
\label{alg:distill_mt}
\begin{algorithmic}
\State $\textbf{Require:}$\\
Replay Buffers $\mathcal{D}^\beta_{k} \equiv \{(\textbf{o}_t, \textbf{g}_t)_{t=0}^{T}\}_{n=0}^N \forall k \in \{1, \cdots, K\}$\\
Trained task-specific  policy $\pi^{k} \forall k \in \{1, \cdots, K\}$\\
Parametric networks for behavior distillation (i.e., the encoder and the policy)\\
\\
\State $\textbf{Algorithm:}$
\State $\Pi \gets \text{randomly initialized}$
\For{each gradient step $t$ = $1...T$}
    \State $k \sim \{1, \cdots, K\}$ \Comment{Sample task}
    \State $\textbf{o}_k, \textbf{g}_k \sim \mathcal{D}^{\beta}_k$
    \State $\textbf{a}_k = \pi^{k}(\textbf{o}_k, \textbf{g}_k)$  
    \State $\hat{\textbf{a}}_k = \Pi(\textbf{o}_k, \textbf{g}_k)$  
    \State $\mathcal{L} = \|\textbf{a}_k - \hat{\textbf{a}}_k\|^{2}$ \Comment{Compute distillation loss (Eq.~\ref{eq:distillation})}
    \State $\text{Update}~\Pi~\text{using loss}~\mathcal{L}$
\EndFor
\end{algorithmic}
\end{algorithm}


\section{Algorithmic Details}
\label{appendix:algorithmic_details}
\method{} utilizes efficient demonstration-guided RL with multi-task distillation to enable both multi-task and lifelong learning. Algorithm~\ref{alg:demoRL} and Algorithm~\ref{alg:distill_mt} describe the algorithm for demonstration-guided RL and behavior distillation for multi-task policy learning. Further implementation details are as follows:

\paragraph{Task expert training procedure} Our model consists of 3 primary neural networks - the encoder, the actor, and the critic. During the BC pretraining phase in Sec.~\ref{subsec:rl}, the encoder and the actor are trained using a mean squared error (MSE) on the expert demonstrations. Next, for finetuning, weights of the pretrained encoder and actor are loaded from memory and the critic is initialized randomly. We observed that the performance of the algorithm is not very sensitive to the value of $\alpha$ (in Eq.~\ref{eq:rl}) and we set it to $0.3$ for all experiments in this paper. Also, the weight $\lambda$ controlling the relative contribution of the two losses in Eq.~\ref{eq:rl} is set to $0.25$ for all the experiments in the paper.

\paragraph{Behavior distillation training procedure} Following the notation used in Sec.~\ref{subsec:distil}, the expert policies $\pi_{1:k}$ are unified into a multitask policy $\Pi$ using Eq.~\ref{eq:distillation}. For lifelong learning, each unified policy $\Pi_{i} \forall i\in \{2, \cdots, N\}$ is obtained by distilling $\pi_{1:i}$ into a single policy using Eq.~\ref{eq:distillation}.

\paragraph{Actor-critic based reward maximization} We use a recent n-step DDPG proposed by \cite{yarats2021mastering} as our RL backbone. The deterministic actor is trained using deterministic policy gradients (DPG) given by Eq.~\ref{eq:actor}. 

\begin{equation}
    \label{eq:actor}
    \begin{aligned}
        \nabla_{\phi} J & \approx \mathbb{E}_{s_{t}\sim\rho_{\beta}} \left[\nabla_{\phi} \left. Q_{\theta}(s,a)\right|_{s=s_{t}, a=\pi_{\phi}(s_{t})} \right]\\
        & = \mathbb{E}_{s_{t}\sim\rho_{\beta}} \left[\nabla_{a} \left. Q_{\theta}(s,a)\right|_{s=s_{t}, a=\pi_{\phi}(s_{t})} \nabla_{\phi} \left. \pi_{\phi}(s)\right|_{s=s_{t}} \right]
    \end{aligned}
\end{equation}

The critic is trained using clipped double Q-learning similar to ~\cite{yarats2021mastering} in order to reduce the overestimation bias in the target value. This is done using two Q-functions, $Q_{\theta1}$ and $Q_{\theta2}$. The critic loss for each critic is given by the equation

\begin{equation}
    \mathcal{L}_{\theta_{k}} = \mathbbm{E}_{(s,a)\sim D_{\beta}} \left[(Q_{\theta_{k}}(s,a) - y)^{2} \right] \forall ~k \in \{1,2\}
\end{equation}

where $\mathcal{D}_{\beta}$ is the replay buffer for online rollouts and $y$ is the target value for n-step DDPG given by

\begin{equation}
    y = \sum_{i=0}^{n-1} \gamma^{i}r_{t+i} + \gamma^{n} \underset{k=1,2}{min}Q_{\bar \theta_{k}}(s_{t+n}, a_{t+n})
\end{equation}

Here, $\gamma$ is the discount factor, $r$ is the reward, and $\bar \theta_{1}$, $\bar \theta_{2}$ are the slow moving weights of target Q-networks. 

\paragraph{Target feature processor to stabilize OT rewards} In Sec.~\ref{subsec:ablations}, we show that in the absence of environment rewards, we can use optimal transport (OT) based trajectory matching rewards for optimizing the task-specific expert during the finetuning phase of Sec.~\ref{subsec:rl}. The OT rewards are computed on the output of the feature processor $f_{\phi}$ which is initialized with a parametric neural network. Hence, as the weights of $f_{\phi}$ change during training, the rewards become non-stationary resulting in unstable training. In order to increase the stability of training, the OT rewards are computed using a target feature processor $f_{\phi^{'}}$~\cite{cohen2022imitation} which is updated with the weights of $f_{\phi}$ every $T_{update}$ environment steps. 

\subsection{Hyperparameters}
The list of hyperparameters for demonstration-guided RL is provided in Table~\ref{table:hyperparams_RL} and for behavior distillation and lifelong learning have been provided in Table~\ref{table:hyperparams_bd_ll}.

\begin{table*}[ht!]
    \begin{center}
    \setlength{\tabcolsep}{18pt}
    \renewcommand{\arraystretch}{1.5}
    \caption{List of hyperparameters for demonstration-guided RL.}
    \label{table:hyperparams_RL}
    \begin{tabular}{ p{4cm} c } 
        \hline
        Parameter & Value \\
        \hline
        Replay buffer size & 150000 \\
        Learning rate      & $1e^{-4}$\\
        Discount $\gamma$   & 0.99\\
        $n$-step returns   & 3 (for Meta-World, DM control),\\
                           & 1 (for Franka kitchen, real robot)\\
        Action repeat      & 2\\
        Seed frames        & 12000\\
        Mini-batch size    & 256\\
        Agent update frequency & 2\\
        Critic soft-update rate & 0.01\\
        Hidden dim         & 1024\\
        Optimizer          & Adam\\
        Exploration steps   & 2000\\
        DDPG exploration schedule & 0.1\\
        Fixed weight $\alpha$       & 0.3\\
        Regularization weight $\lambda$ & 0.25\\
        Target feature processor update frequency(steps) for OT & 20000\\
        Reward scale factor for OT & 10\\
        \hline
    \end{tabular}
    \end{center}
\end{table*}

\begin{table*}[ht!]
    \begin{center}
    \setlength{\tabcolsep}{18pt}
    \renewcommand{\arraystretch}{1.5}
    \caption{List of hyperparameters for multi-task learning and lifelong learning using Behavior Distillation.}
    \label{table:hyperparams_bd_ll}
    \begin{tabular}{ p{6cm} c } 
        \hline
        Parameter & Value \\
        \hline
        Replay buffer size for each task & 30000 \\
        Learning rate      & $5e^{-5}$\\
        Action repeat      & 2\\
        Mini-batch size    & 256\\
        Hidden dim         & 1024\\
        Optimizer          & Adam\\
        \hline
    \end{tabular}
    \end{center}
\end{table*}


\begin{table*}[t!]
    \begin{center}
    \setlength{\tabcolsep}{5pt}
    \renewcommand{\arraystretch}{1.5}
    \caption{List of tasks used for evaluation.}
    \label{table:tasks}
    \begin{tabular}{ c c c c } 
     \hline
     Suite & Tasks & Allowed Steps & \# Demonstrations \\
     \hline
     DeepMind Control     &  Cheetah Run           & $5\times10^{5}$ & 10 \\
     \hline
     Meta-World           & Door Unlock            & $5\times10^{4}$   & 1\\ 
                          & Door Open              & $1\times10^{5}$   &\\
                          & Plate Slide            & $5\times10^{4}$   &\\
                          & Plate Slide Back       & $5\times10^{4}$   &\\
                          & Plate Slide Side       & $5\times10^{4}$   &\\
                          & Plate Slide Side Back  & $5\times10^{4}$   &\\
                          & Coffee Button          & $5\times10^{4}$   &\\
                          & Coffee Pull            & $1.5\times10^{5}$ &\\
                          & Button Press           & $5\times10^{4}$   &\\
                          & Button Press Wall      & $5\times10^{4}$   &\\
                          & Button Press Topdown   & $1\times10^{5}$   &\\
                          & Button Press Wall Topdown & $1.5\times10^{5}$ &\\
                          & Bin Picking            & $1.5\times10^{5}$ &\\
                          & Hammer                 & $1\times10^{5}$   &\\
                          & Drawer Close           & $5\times10^{4}$   &\\
                          & Drawer Open            & $5\times10^{4}$   &\\
     \hline
     Franka Kitchen       & Bottom Burner          & $4\times10^{5}$ & 100\\ 
                          & Top Burner             &                 & \\
                          & Light Switch           &                 & \\
                          & Slide Cabinet          &                 & \\
                          & Microwave              &                 & \\
                          & Kettle                 &                 & \\
     \hline
     xArm Robot           & Put Peg in Green Cup   & $6\times10^{3}$ & 2\\ 
                          & Open a Box             &                 & 1\\
                          & Pouring Grains         &                 & 1\\
                          & Put Peg in Yellow Cup  &                 & 2\\
                          & Reach                  &                 & 1\\
                          & Drawer Close           &                 & 1\\
    \hline
    \end{tabular}
    \end{center}
\end{table*}

\section{Environments}
\label{appendix:envs}
\begin{figure*}[t]
    \centering
    \includegraphics[width=0.9\linewidth]{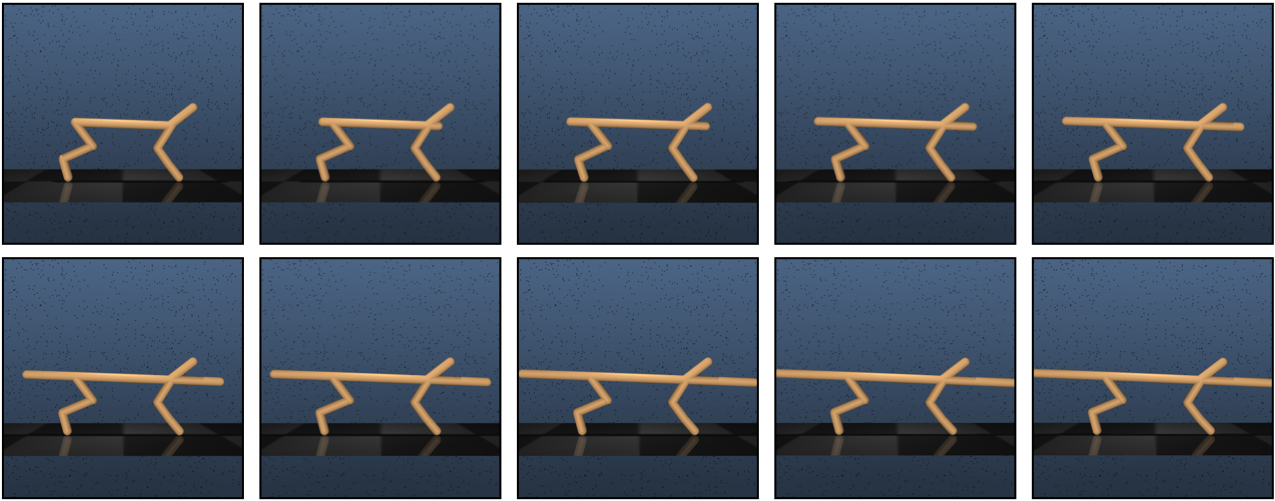}
    \caption{A visualization of variations of the tasks in the cheetah run environment in the DeepMind Control suite. The torso length of the cheetah is varied across tasks.}
    \label{fig:dmc_envs}
\end{figure*}

\begin{figure*}[t]
    \centering
    \includegraphics[width=0.8\linewidth]{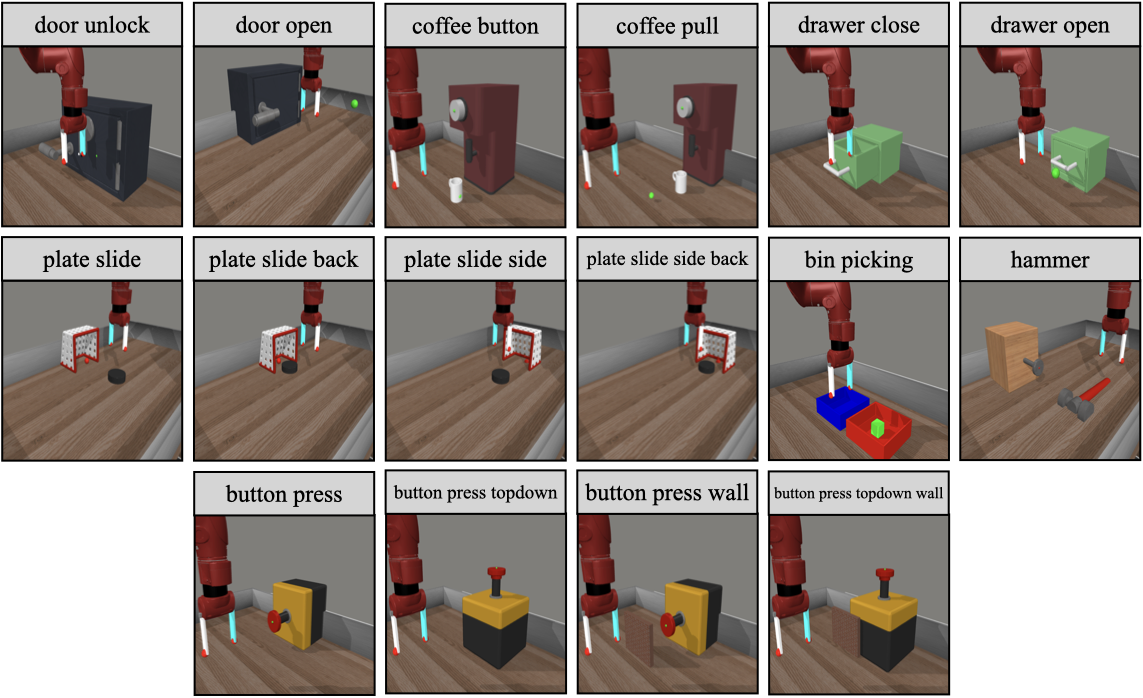}
    \caption{A visualization of the 16 different tasks from the Meta-World suite used for our experiments.}
    \label{fig:mw_envs}
\end{figure*}

\begin{figure*}[t]
    \centering
    \includegraphics[width=0.7\linewidth]{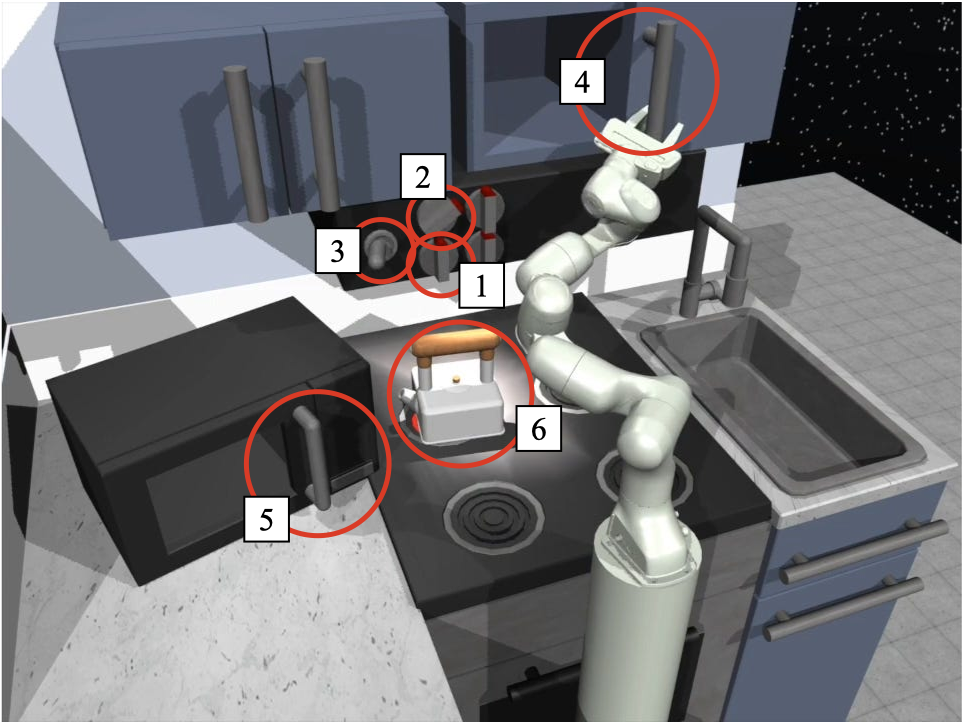}
    \caption{A visualization of the Franka Kitchen environment. We experiment with 6 tasks in this environment - $(1)$ activate bottom burner, $(2)$ activate top burner, $(3)$ turn on the light switch, $(4)$ open the slide cabinet, $(5)$ open the microwave door, and $(6)$ move the kettle to the top left burner.}
    \label{fig:kitchen_tasks}
\end{figure*}

Table~\ref{table:tasks} lists the different tasks that we experiment with from the DeepMind Control suite~\cite{tassa2018deepmind,todorov2012mujoco}, the Meta-World suite~\cite{yu2019meta} and the Franka kitchen environment~\cite{gupta2019relay} along with the number of training steps and the number of demonstrations used. The episode length for all tasks in DeepMind Control is 1000 steps, for Meta-world is 125 steps (except bin picking which runs for 175 steps), and for Franka kitchen is 280 steps. A visualization of all the tasks in DeepMind Control, Meta-World, and Franka kitchen has been provided in Fig.~\ref{fig:dmc_envs}, Fig.~\ref{fig:mw_envs} and Fig.~\ref{fig:kitchen_tasks} respectively.

\section{Demonstrations}
\label{appendix:demos}
The details about collecting demonstrations on the 3 simulated environment suites have been mentioned in Sec.~\ref{subsec:experimental_setup}. Here we provide some additional details about splitting the 566 demonstrations for the Franka kitchen environment provided in the relay policy learning dataset~\cite{gupta2019relay} into task-specific snippets. Each demonstration carries out 4 tasks in the environment. We roll out each demonstration in the environment to figure out the time step at which a task is completed and record the task that was completed. After doing this for all demonstrations, we divide each trajectory into task-specific snippets by considering the frames from after the previous task was completed till the current one is completed. Each such snippet is appended to the task-specific demonstration set. We reiterate that since this is play data, the task-specific snippets are suboptimal in the sense that each snippet may start from where the previous task was completed. These positions might be very different from where the Franka robot arm is initialized in the environment when we want to perform a specific task.

For the real robot experiments, up to 2 demonstrations are collected per task by a human operator using a joystick.


\begin{figure*}[t!]
    \centering
    \includegraphics[width=0.8\linewidth]{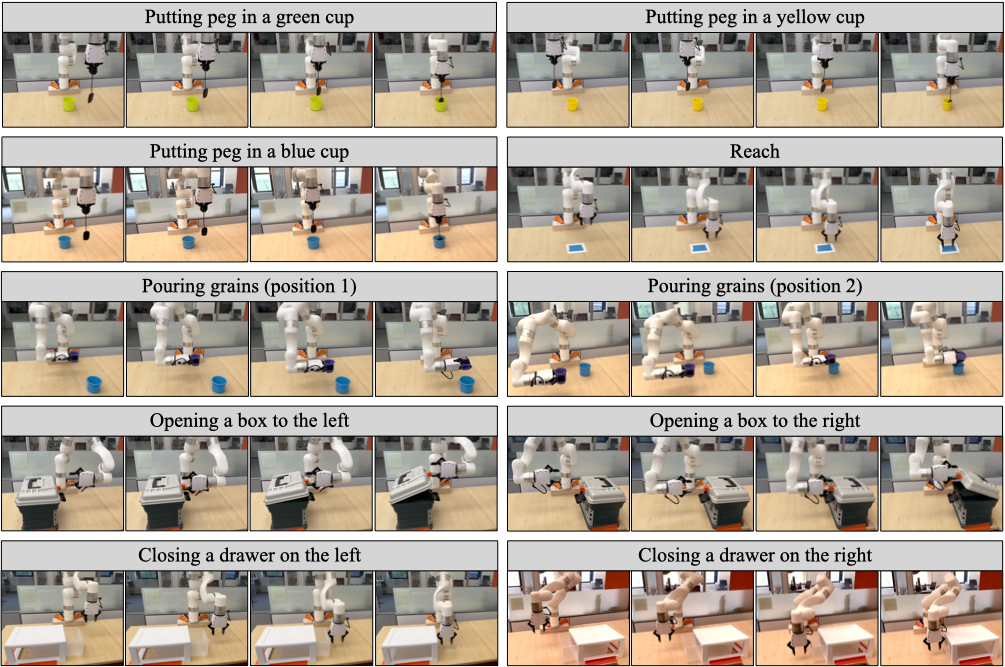}    
    \caption{A visualization of trajectory rollouts of the 10 real-world tasks used for evaluations.}
    \label{appendix:fig:real_world_tasks}
\end{figure*}

\section{Robot tasks}
\label{appendix:robot_tasks}
We evaluate \method{} on 10 real-world tasks by adding 4 variations of tasks in addition to the 6 real-world tasks depicted in Fig.~\ref{fig:intro}. In this section, we describe the suite of manipulation experiments carried out on a real robot in this paper. A visualization of a trajectory rollout of each task has been provided in Fig.~\ref{appendix:fig:real_world_tasks}.

\begin{enumerate}[label=(\alph*),leftmargin=*]
    \item \textbf{Insert peg in green cup:} The robot arm is supposed to insert a peg, hanging by a string, into a green cup placed on the table.
    \item \textbf{Insert peg in yellow cup:} The robot arm is supposed to insert a peg, hanging by a string, into a yellow cup placed on the table.
    \item \textbf{Insert peg in blue cup:} The robot arm is supposed to insert a peg, hanging by a string, into a blue cup placed on the table.
    \item \textbf{Reach:} The robot arm is required to reach a specific goal after being initialized at a random position.
    \item \textbf{Pouring grains (position 1):} Given a cup with some item placed inside (in our case, grains), the robot arm is supposed to move towards a cup placed on the table and pour the item into the cup.
    \item \textbf{Pouring grains (position 2):} Given a cup with some item placed inside (in our case, grains), the robot arm is supposed to move towards a cup placed on the table and pour the item into the cup.
    \item \textbf{Open a box (left):} In this task, the robot arm is supposed to open the lid of a box placed on the left of the table by lifting a handle provided in the front of the box.
    \item \textbf{Open a box (right):} In this task, the robot arm is supposed to open the lid of a box placed on the right of the table by lifting a handle provided in the front of the box.
    \item \textbf{Drawer Close (left):} Here, the robot arm is supposed to close an open drawer placed towards the left of the table by pushing it to the target.    
    \item \textbf{Drawer Close (right):} Here, the robot arm is supposed to close an open drawer placed towards the right of the table by pushing it to the target.    
\end{enumerate}

\paragraph{Evaluation procedure} For each task, we obtained a set of 10 random initializations and evaluate 4 methods (shown in Table~\ref{table:real_results}, namely single task BC, single task expert, and \method{} in a multi-task and lifelong setting) over 10 trajectories from the same set of initializations. These initializations are different for each task based on the limits of the observation space for the task.


\section{Baseline methods}
\label{appendix:baselines}
Throughout the paper, we compare \method{} with several prominent imitation learning and reinforcement learning methods. Here, we give a brief description of each of the baseline models that have been used.

\begin{enumerate}[label=(\alph*),leftmargin=*]
    \item \textbf{Task-specific expert:} In DeepMind Control, 10 expert demonstrations per task-variant are collected by training expert policies using DrQ-v2 (as mentioned in Sec. 3.1). These demonstrations are used to learn task-specific expert policies using demonstration-guided RL. Similarly, the demonstrations collected using the scripted policies for Meta-World are then used to learn task-specific expert policies using demonstration-guided RL. For Franka Kitchen, as mentioned in Section 3.1, we use an offline dataset comprising 566 demonstrations. However, since these demonstrations comprise randomly ordered play data, we segment each trajectory into task-specific snippets and use 100 demonstrations per task to learn task-specific experts using demonstration-guided RL. These task-specific expert policies are the highest-performing policies that we are able to learn from the collected demonstrations and online demonstration-guided RL. The specialist policies are unified into a generalist agent through knowledge distillation to obtain the PolyTask policy.
    \item \textbf{Single-task behavior cloning (BC):} A supervised learning framework where a single-task policy $\pi(\cdot| o)$ is learned given a dataset of (observation, action) tuples $(o, a)$.
    \item \textbf{Goal conditioned BC (GCBC)}~\cite{lynch2020learning,emmons2021rvs}: A supervised learning framework where a goal conditioned multi-task policy $\pi(\cdot| o, g)$ is learned given a dataset of (observation, action, goal) tuples $(o, a, g)$.
    \item \textbf{Multi-task RL (MTRL)}~\cite{vithayathil2020survey,teh2017distral,sodhani2021multi, kalashnikov2021mt}: A framework where a policy $\pi(\cdot| o, g)$ is jointly learned on all tasks or environments using reinforcement learning. In MTRL, the agent is assumed to have access to all tasks, or environments simultaneously. Here, we consider a DrQ-v2 policy~\cite{yarats2021mastering} and jointly optimize it across multiple tasks by randomly sampling a task for each environment rollout and learning a goal-conditioned policy on randomly sample training samples across all tasks from a replay buffer.
    \item \textbf{Distral}~\cite{teh2017distral}: A MTRL algorithm that jointly trains separate task-specific policies and a distilled policy that captures common behavior across tasks.\SH{Explain}
    \item \textbf{MTRL-PCGrad}~\cite{yu2020gradient}: A variant of MTRL with {projecting-conflicting-gradients (PCGrad)} style gradient optimization to mitigate task interference during gradient updates. PCGrad requires computing the loss on all tasks before each gradient update which increases the time per gradient update. This is why we could not get the result for MTRL-PCGrad on the Franka kitchen environment in Table~\ref{table:mt_results}.
    \item \textbf{MTRL-Demo}: A demo-guided variant of MTRL where we adapt the strategy proposed in Sec.~\ref{subsec:rl} to a multi-task setting. A goal-conditioned multi-task BC policy is first trained jointly on all task demonstrations and this policy is used for initialization and regularization during the MTRL training. Though MTRL performs poorly on the harder tasks in the Meta-World suite and Franka kitchen environment in Table~\ref{table:mt_results}, we observe that MTRL-demo does significantly better thus highlighting the importance of demonstration-based guidance for policy learning using RL.
    \item \textbf{Fine-tuning}~\cite{julian2020efficient}: A framework where a single policy is initialized and keeps getting finetuned on each new task. The parameter count of the policy remains constant throughout training. Even if such a policy does well on the most recent task, it tends to forget previous tasks when being trained on future ones.
    \item \textbf{Progressive Nets}~\cite{rusu2016progressive}: A framework that deals with catastrophic forgetting by adding a new model for each task with lateral connections to all previous models to allow for forward transfer. The parameter count of the policy increases with each new task. In this work, we use progressive networks in an RL setting as done in prior work~\cite{rusu2016progressive,xie2022lifelong}.
\end{enumerate}


\section{Behavior Distillation}
\label{appendix:behavior_distillation}
In Sec.~\ref{subsec:distil}, we introduce behavior distillation, a technique used for combining multiple expert policies into a single unified policy with the use of knowledge distillation. During the process, we randomly initialize a policy $\Pi$ having the same architecture as the expert policies. We randomly sample observations $o$ and corresponding goals $g$ from the replay buffer for each expert policy $\pi_k$, compute the corresponding expert actions $\pi_{k}(o, g)$ and use $(s, \pi_{k}(o, g))$ to train $\Pi$ using the mean squared error (MSE) loss shown in Eq.~\ref{eq:distillation}. It must be noted that though we initialize the distillation policy $\Pi$ to have the same architecture as the task expert, our framework is compatible with having a different architecture and model size for the distilled policy and the expert policies. The hyperparameters used for multi-task policy learning using behavior distillation have been provided in Table~\ref{table:hyperparams_bd_ll}.


\section{Lifelong Learning}
\label{appendix:lifelong_polytask}
\begin{figure*}[t]
    \centering
    \includegraphics[width=0.7\linewidth]{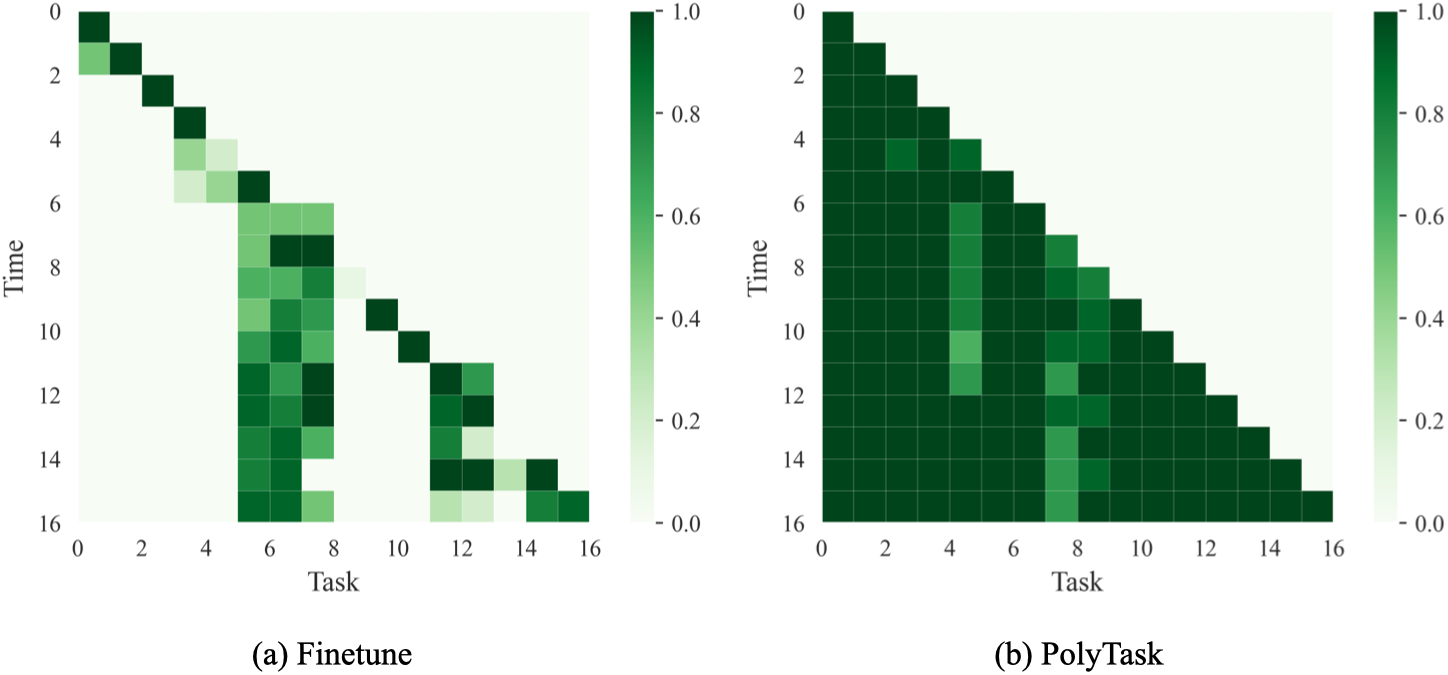}
    \caption{A comparison between the performance of Fine-tuning and \method{} on the Meta-World suite in a lifelong learning setting. We observe that \method{} exhibits a significantly better ability to tackle catastrophic forgetting. \SH{Change figure}}
    \label{fig:metaworld_finetune_lifelong}
\end{figure*}

\begin{figure*}[t]
    \centering
    \includegraphics[width=0.7\linewidth]{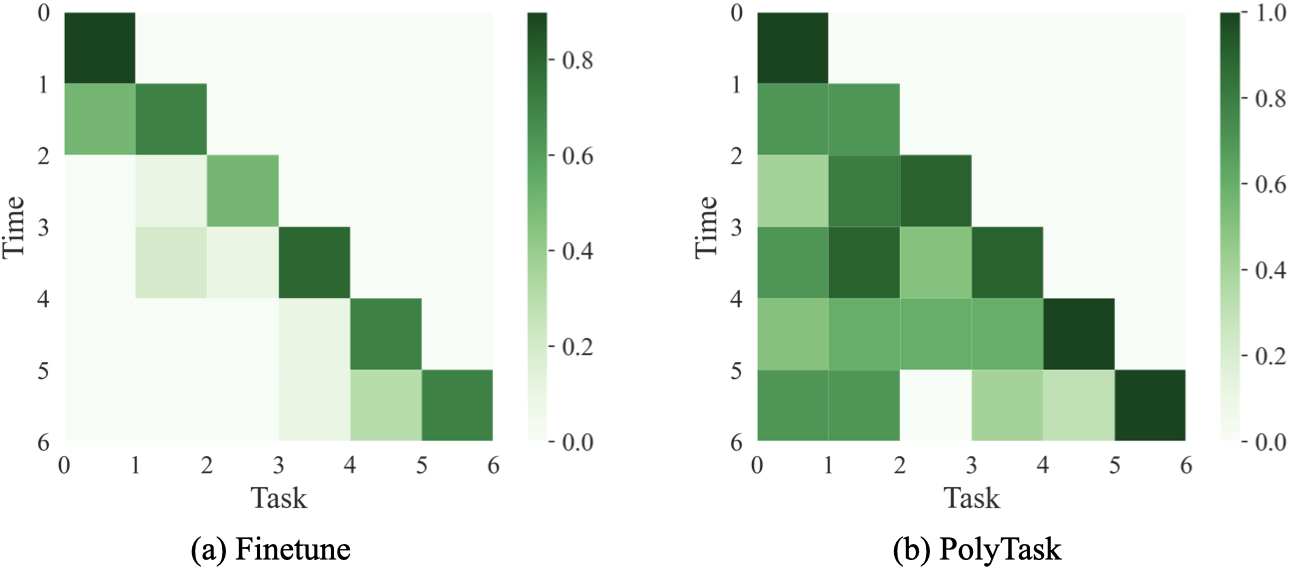}
    \caption{A comparison between the performance of Fine-tuning and \method{} on the Franka kitchen environment in a lifelong learning setting. We observe that \method{} exhibits a significantly better ability to tackle catastrophic forgetting.}
    \label{fig:kitchen_finetune_lifelong}
\end{figure*}

In this work, we claim that the offline training procedure of behavior distillation makes it suitable for lifelong learning.
Fig.~\ref{fig:metaworld_finetune_lifelong} and Fig.~\ref{fig:kitchen_finetune_lifelong} show a comparison between two of our baselines - finetune and \method{} - in a lifelong learning scenario on the Meta-World suite and the Franka kitchen environment respectively. We observe that \method{} exhibits a significantly better ability to tackle catastrophic forgetting than just finetuning the policy. The hyperparameters used for lifelong learning using behavior distillation have been provided in Table~\ref{table:hyperparams_bd_ll}. 

\section{Additional Analysis and Ablations}
\label{appendix:learning}

\begin{table*}[t]
\centering
\caption{\method{} results for multi-task learning on 10 real-world tasks.}
\label{appendix:table:real_results}
\begin{tabular}{lccc}
\hline
\multicolumn{1}{c}{\textbf{Task}} & \textbf{Single-task BC} & \textbf{Single-task expert} & \textbf{Polytask} \\ \hline
Insert Peg in Green Cup           & 4/10                    & 5/10                        & 7/10              \\
Insert Peg in Yellow Cup          & 5/10                    & 9/10                        & 10/10             \\
Insert Peg in Blue Cup            & 4/10                    & 10/10                       & 10/10                  \\
Reach                             & 3/10                    & 10/10                       & 10/10             \\
Pouring grains (position 1)       & 3/10                    & 6/10                        & 6/10              \\
Pouring grains (position 2)       & 3/10                    & 9/10                        & 9/10                 \\
Open box (left)                   & 0/10                    & 10/10                       & 10/10             \\
Open box (right)                  & 1/10                    & 10/10                       & 9/10                 \\
Drawer Close (left)               & 5/10                    & 10/10                       & 10/10             \\
Drawer Close (right)              & 4/10                    & 10/10                       & 10/10        \\\hline        
\end{tabular}
\end{table*}

\subsection{Multi-task distillation on real-robot tasks}
In addition to results shown in Sec.~\ref{subsec:real_robot}, Table~\ref{appendix:table:real_results} depicts the performance of \method{} for multi-task policy learning on the suite of 10 real-world tasks. We observe that the multi-task policy learned using \method{} is able to effectively succeed on 9.1 out of 10 tasks and outperforms the single-task experts which effectively succeed on 8.9 out of 10 tasks. This highlights the potential advantage of training a single policy for multiple tasks.

\begin{figure*}[t!]
    \centering
    \includegraphics[width=0.7\linewidth]{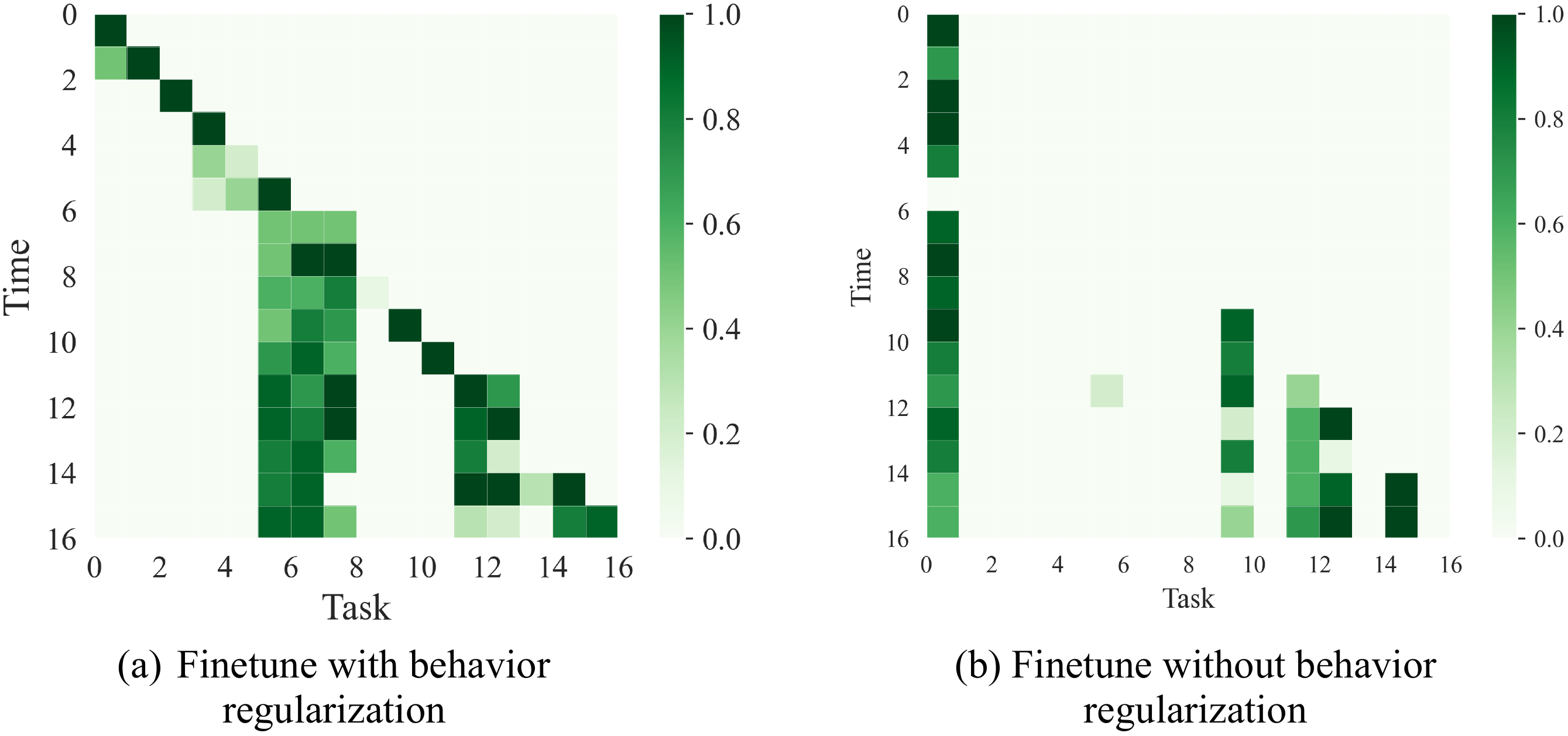}
    \caption{An illustration of the importance of behavior regularization for training a multi-task policy on the Meta-World suite when the tasks are presented sequentially. We observe that without behavior regularization, most of the skills are not learned within the given interaction budget.}
    \label{appendix:fig:mw_importance_of_bcreg}
\end{figure*}

\subsection{How important is adding BC regularization?}
In Sec.~\ref{subsec:ll_exps}, we add a behavior regularization term to the learning objective while training finetuning and progressive networks. Behavior regularization is aimed at accelerating the learning of the current task. Fig.~\ref{appendix:fig:mw_importance_of_bcreg} justifies this choice by highlighting the importance of behavior regularization in learning skills sequentially using a finetuning network. We observe that without behavior regularization, most of the skills are not learned within the given interaction budget. However, in the absence of behavior regularization, once a skill is learned, the policy tends to remember it. This can be attributed to the fact that the policy is always trained on replay buffer data from all previous tasks without any interference from the regularization loss. Thus, to summarize, behavior regularization can enhance the sample efficiency of RL at the cost of forgetting prior experiences.

\begin{figure*}[t!]
    \centering
    \includegraphics[width=0.9\linewidth]{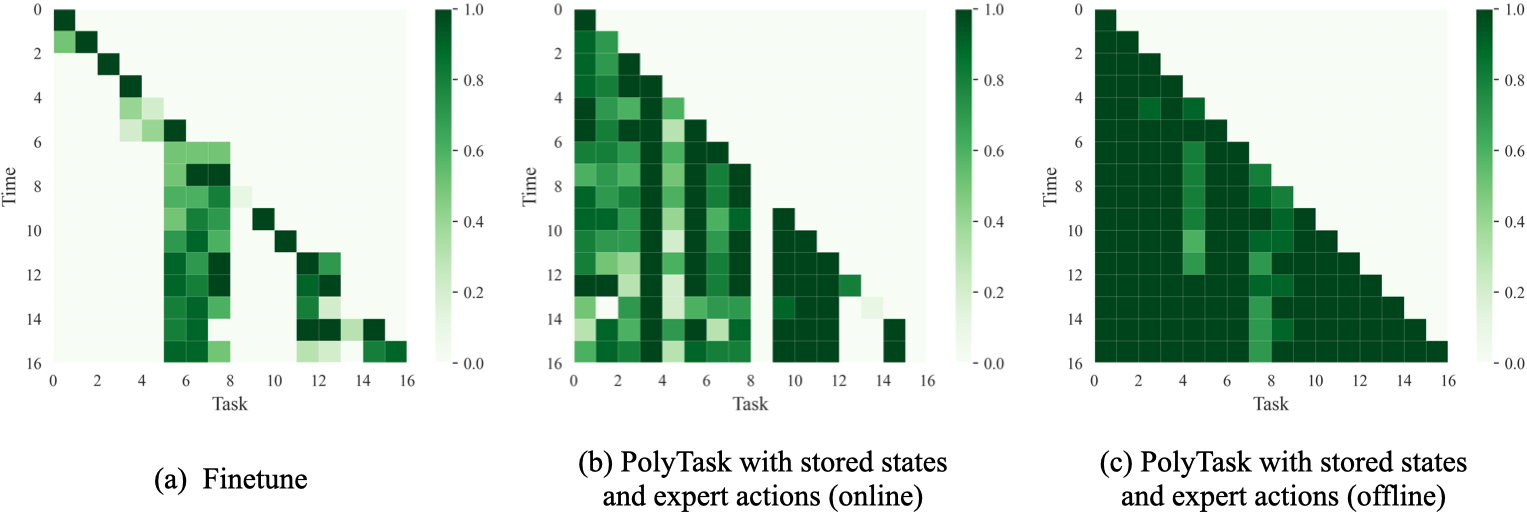}
    \caption{An illustration of the performance of \method{} in a lifelong learning scenario on the Meta-World suite when utilizing relabeled expert actions in the task-specific replay buffers. We observe that though there's an extra cost of iterating through the entire replay buffer and storing expert actions, the described variants can significantly reduce catastrophic forgetting.\SH{Only keep A, C, D}}
    \label{appendix:fig:variant_ll}
\end{figure*}

\subsection{Variants of \method{} for lifelong learning}
In this section, we study two variants of distillation strategies for \method{}. We study two variants of the distillation procedure - \textit{online} and \textit{offline}. A description of the variants has been provided below:

\paragraph{Online} In this variant, the policy is trained in a similar way as fine-tuning. However, at the end of each task, the actions in the replay buffer corresponding to the task are relabeled using the current lifelong policy. Then while training on a new task, a distillation loss is used for all previous tasks while an RL is used loss for the new task. The performance of the described method has been shown in Fig.~\ref{appendix:fig:variant_ll}(b).

\paragraph{Offline} In this variant, a single-task expert policy is obtained for each task using demonstration-guided RL (Eq.~\ref{eq:rl}). At the end of each task, the actions in the replay buffer corresponding to the task are relabeled using the task-specific expert policy. Then the lifelong learning policy $\Pi_N$ is learned by distilling the task experts (using the task-specific state and expert action tuples) into a single policy. The performance of the described method has been shown in Fig.~\ref{appendix:fig:variant_ll}(c).

From the results provided in Fig.~\ref{appendix:fig:variant_ll}, we conclude that both variants of \method{} significantly reduce catastrophic forgetting lifelong learning as compared to finetuning. Further, the offline variant of \method{} exhibits better performance than its online counterpart.

%% file: root.bbl
\begin{thebibliography}{10}
\providecommand{\url}[1]{#1}
\csname url@rmstyle\endcsname
\providecommand{\newblock}{\relax}
\providecommand{\bibinfo}[2]{#2}
\providecommand\BIBentrySTDinterwordspacing{\spaceskip=0pt\relax}
\providecommand\BIBentryALTinterwordstretchfactor{4}
\providecommand\BIBentryALTinterwordspacing{\spaceskip=\fontdimen2\font plus
\BIBentryALTinterwordstretchfactor\fontdimen3\font minus \fontdimen4\font\relax}
\providecommand\BIBforeignlanguage[2]{{%
\expandafter\ifx\csname l@#1\endcsname\relax
\typeout{** WARNING: IEEEtran.bst: No hyphenation pattern has been}%
\typeout{** loaded for the language `#1'. Using the pattern for}%
\typeout{** the default language instead.}%
\else
\language=\csname l@#1\endcsname
\fi
#2}}

\bibitem{reed2022generalist}
S.~Reed, K.~Zolna, E.~Parisotto, S.~G. Colmenarejo, A.~Novikov, G.~Barth-Maron, M.~Gimenez, Y.~Sulsky, J.~Kay, J.~T. Springenberg, \emph{et~al.}, ``A generalist agent,'' \emph{arXiv preprint arXiv:2205.06175}, 2022.

\bibitem{driess2023palm}
D.~Driess, F.~Xia, M.~S. Sajjadi, C.~Lynch, A.~Chowdhery, B.~Ichter, A.~Wahid, J.~Tompson, Q.~Vuong, T.~Yu, \emph{et~al.}, ``Palm-e: An embodied multimodal language model,'' \emph{arXiv preprint arXiv:2303.03378}, 2023.

\bibitem{shridhar2023perceiver}
M.~Shridhar, L.~Manuelli, and D.~Fox, ``Perceiver-actor: A multi-task transformer for robotic manipulation,'' in \emph{Conference on Robot Learning}.\hskip 1em plus 0.5em minus 0.4em\relax PMLR, 2023, pp. 785--799.

\bibitem{brohan2022rt}
A.~Brohan, N.~Brown, J.~Carbajal, Y.~Chebotar, J.~Dabis, C.~Finn, K.~Gopalakrishnan, K.~Hausman, A.~Herzog, J.~Hsu, \emph{et~al.}, ``Rt-1: Robotics transformer for real-world control at scale,'' \emph{arXiv preprint arXiv:2212.06817}, 2022.

\bibitem{lu2022unified}
J.~Lu, C.~Clark, R.~Zellers, R.~Mottaghi, and A.~Kembhavi, ``Unified-io: A unified model for vision, language, and multi-modal tasks,'' \emph{arXiv preprint arXiv:2206.08916}, 2022.

\bibitem{kolesnikov2022uvim}
A.~Kolesnikov, A.~Susano~Pinto, L.~Beyer, X.~Zhai, J.~Harmsen, and N.~Houlsby, ``Uvim: A unified modeling approach for vision with learned guiding codes,'' \emph{Advances in Neural Information Processing Systems}, vol.~35, pp. 26\,295--26\,308, 2022.

\bibitem{Casas_2021_CVPR}
S.~Casas, A.~Sadat, and R.~Urtasun, ``Mp3: A unified model to map, perceive, predict and plan,'' in \emph{Proceedings of the IEEE/CVF Conference on Computer Vision and Pattern Recognition (CVPR)}, June 2021, pp. 14\,403--14\,412.

\bibitem{dong2019unified}
L.~Dong, N.~Yang, W.~Wang, F.~Wei, X.~Liu, Y.~Wang, J.~Gao, M.~Zhou, and H.-W. Hon, ``Unified language model pre-training for natural language understanding and generation,'' \emph{Advances in neural information processing systems}, vol.~32, 2019.

\bibitem{girdhar2023imagebind}
R.~Girdhar, A.~El-Nouby, Z.~Liu, M.~Singh, K.~V. Alwala, A.~Joulin, and I.~Misra, ``Imagebind: One embedding space to bind them all,'' \emph{arXiv preprint arXiv:2305.05665}, 2023.

\bibitem{bao2020unilmv2}
H.~Bao, L.~Dong, F.~Wei, W.~Wang, N.~Yang, X.~Liu, Y.~Wang, J.~Gao, S.~Piao, M.~Zhou, \emph{et~al.}, ``Unilmv2: Pseudo-masked language models for unified language model pre-training,'' in \emph{International conference on machine learning}.\hskip 1em plus 0.5em minus 0.4em\relax PMLR, 2020, pp. 642--652.

\bibitem{uchendu2022jump}
I.~Uchendu, T.~Xiao, Y.~Lu, B.~Zhu, M.~Yan, J.~Simon, M.~Bennice, C.~Fu, C.~Ma, J.~Jiao, \emph{et~al.}, ``Jump-start reinforcement learning,'' \emph{arXiv preprint arXiv:2204.02372}, 2022.

\bibitem{jena2020augmenting}
R.~Jena, C.~Liu, and K.~Sycara, ``Augmenting gail with bc for sample efficient imitation learning,'' \emph{arXiv preprint arXiv:2001.07798}, 2020.

\bibitem{haldar2023watch}
S.~Haldar, V.~Mathur, D.~Yarats, and L.~Pinto, ``Watch and match: Supercharging imitation with regularized optimal transport,'' in \emph{Conference on Robot Learning}.\hskip 1em plus 0.5em minus 0.4em\relax PMLR, 2023, pp. 32--43.

\bibitem{haldar2023teach}
S.~Haldar, J.~Pari, A.~Rai, and L.~Pinto, ``Teach a robot to fish: Versatile imitation from one minute of demonstrations,'' \emph{arXiv preprint arXiv:2303.01497}, 2023.

\bibitem{smith2022walk}
L.~Smith, I.~Kostrikov, and S.~Levine, ``A walk in the park: Learning to walk in 20 minutes with model-free reinforcement learning,'' \emph{arXiv preprint arXiv:2208.07860}, 2022.

\bibitem{ball2023efficient}
P.~J. Ball, L.~Smith, I.~Kostrikov, and S.~Levine, ``Efficient online reinforcement learning with offline data,'' \emph{arXiv preprint arXiv:2302.02948}, 2023.

\bibitem{sodhani2021multi}
S.~Sodhani, A.~Zhang, and J.~Pineau, ``Multi-task reinforcement learning with context-based representations,'' in \emph{International Conference on Machine Learning}.\hskip 1em plus 0.5em minus 0.4em\relax PMLR, 2021, pp. 9767--9779.

\bibitem{yu2020gradient}
T.~Yu, S.~Kumar, A.~Gupta, S.~Levine, K.~Hausman, and C.~Finn, ``Gradient surgery for multi-task learning,'' \emph{Advances in Neural Information Processing Systems}, vol.~33, pp. 5824--5836, 2020.

\bibitem{teh2017distral}
Y.~Teh, V.~Bapst, W.~M. Czarnecki, J.~Quan, J.~Kirkpatrick, R.~Hadsell, N.~Heess, and R.~Pascanu, ``Distral: Robust multitask reinforcement learning,'' \emph{Advances in neural information processing systems}, vol.~30, 2017.

\bibitem{vithayathil2020survey}
N.~Vithayathil~Varghese and Q.~H. Mahmoud, ``A survey of multi-task deep reinforcement learning,'' \emph{Electronics}, vol.~9, no.~9, p. 1363, 2020.

\bibitem{robins1993catastrophic}
A.~Robins, ``Catastrophic forgetting in neural networks: the role of rehearsal mechanisms,'' in \emph{Proceedings 1993 The First New Zealand International Two-Stream Conference on Artificial Neural Networks and Expert Systems}.\hskip 1em plus 0.5em minus 0.4em\relax IEEE, 1993, pp. 65--68.

\bibitem{rusu2015policy}
A.~A. Rusu, S.~G. Colmenarejo, C.~Gulcehre, G.~Desjardins, J.~Kirkpatrick, R.~Pascanu, V.~Mnih, K.~Kavukcuoglu, and R.~Hadsell, ``Policy distillation,'' \emph{arXiv preprint arXiv:1511.06295}, 2015.

\bibitem{julian2020efficient}
R.~Julian, B.~Swanson, G.~S. Sukhatme, S.~Levine, C.~Finn, and K.~Hausman, ``Efficient adaptation for end-to-end vision-based robotic manipulation,'' in \emph{4th Lifelong Machine Learning Workshop at ICML 2020}, 2020.

\bibitem{ying2018transfer}
W.~Ying, Y.~Zhang, J.~Huang, and Q.~Yang, ``Transfer learning via learning to transfer,'' in \emph{International conference on machine learning}.\hskip 1em plus 0.5em minus 0.4em\relax PMLR, 2018, pp. 5085--5094.

\bibitem{rusu2016progressive}
A.~A. Rusu, N.~C. Rabinowitz, G.~Desjardins, H.~Soyer, J.~Kirkpatrick, K.~Kavukcuoglu, R.~Pascanu, and R.~Hadsell, ``Progressive neural networks,'' \emph{arXiv preprint arXiv:1606.04671}, 2016.

\bibitem{parisi2017lifelong}
G.~I. Parisi, J.~Tani, C.~Weber, and S.~Wermter, ``Lifelong learning of human actions with deep neural network self-organization,'' \emph{Neural Networks}, vol.~96, pp. 137--149, 2017.

\bibitem{lynch2020learning}
C.~Lynch, M.~Khansari, T.~Xiao, V.~Kumar, J.~Tompson, S.~Levine, and P.~Sermanet, ``Learning latent plans from play,'' in \emph{Conference on robot learning}.\hskip 1em plus 0.5em minus 0.4em\relax PMLR, 2020, pp. 1113--1132.

\bibitem{emmons2021rvs}
S.~Emmons, B.~Eysenbach, I.~Kostrikov, and S.~Levine, ``Rvs: What is essential for offline rl via supervised learning?'' \emph{arXiv preprint arXiv:2112.10751}, 2021.

\bibitem{yarats2021mastering}
D.~Yarats, R.~Fergus, A.~Lazaric, and L.~Pinto, ``Mastering visual continuous control: Improved data-augmented reinforcement learning,'' \emph{arXiv preprint arXiv:2107.09645}, 2021.

\bibitem{hoshino2022opirl}
H.~Hoshino, K.~Ota, A.~Kanezaki, and R.~Yokota, ``Opirl: Sample efficient off-policy inverse reinforcement learning via distribution matching,'' in \emph{2022 International Conference on Robotics and Automation (ICRA)}.\hskip 1em plus 0.5em minus 0.4em\relax IEEE, 2022, pp. 448--454.

\bibitem{gou2021knowledge}
J.~Gou, B.~Yu, S.~J. Maybank, and D.~Tao, ``Knowledge distillation: A survey,'' \emph{International Journal of Computer Vision}, vol. 129, pp. 1789--1819, 2021.

\bibitem{qu2021recent}
H.~Qu, H.~Rahmani, L.~Xu, B.~Williams, and J.~Liu, ``Recent advances of continual learning in computer vision: An overview,'' \emph{arXiv preprint arXiv:2109.11369}, 2021.

\bibitem{xu2022survey}
C.~Xu and J.~McAuley, ``A survey on model compression for natural language processing,'' \emph{arXiv preprint arXiv:2202.07105}, 2022.

\bibitem{lai2020dual}
K.-H. Lai, D.~Zha, Y.~Li, and X.~Hu, ``Dual policy distillation,'' \emph{arXiv preprint arXiv:2006.04061}, 2020.

\bibitem{czarnecki2019distilling}
W.~M. Czarnecki, R.~Pascanu, S.~Osindero, S.~Jayakumar, G.~Swirszcz, and M.~Jaderberg, ``Distilling policy distillation,'' in \emph{The 22nd international conference on artificial intelligence and statistics}.\hskip 1em plus 0.5em minus 0.4em\relax PMLR, 2019, pp. 1331--1340.

\bibitem{ryu2019caql}
M.~Ryu, Y.~Chow, R.~Anderson, C.~Tjandraatmadja, and C.~Boutilier, ``Caql: Continuous action q-learning,'' \emph{arXiv preprint arXiv:1909.12397}, 2019.

\bibitem{tassa2018deepmind}
Y.~Tassa, Y.~Doron, A.~Muldal, T.~Erez, Y.~Li, D.~d.~L. Casas, D.~Budden, A.~Abdolmaleki, J.~Merel, A.~Lefrancq, \emph{et~al.}, ``Deepmind control suite,'' \emph{arXiv preprint arXiv:1801.00690}, 2018.

\bibitem{todorov2012mujoco}
E.~Todorov, T.~Erez, and Y.~Tassa, ``Mujoco: A physics engine for model-based control,'' in \emph{2012 IEEE/RSJ international conference on intelligent robots and systems}.\hskip 1em plus 0.5em minus 0.4em\relax IEEE, 2012, pp. 5026--5033.

\bibitem{Sodhani2021MTEnv}
\BIBentryALTinterwordspacing
S.~Sodhani, L.~Denoyer, P.-A. Kamienny, and O.~Delalleau, ``Mtenv - environment interface for mulit-task reinforcement learning,'' Github, 2021. [Online]. Available: \url{https://github.com/facebookresearch/mtenv}
\BIBentrySTDinterwordspacing

\bibitem{vaswani2017attention}
A.~Vaswani, N.~Shazeer, N.~Parmar, J.~Uszkoreit, L.~Jones, A.~N. Gomez, {\L}.~Kaiser, and I.~Polosukhin, ``Attention is all you need,'' \emph{Advances in neural information processing systems}, vol.~30, 2017.

\bibitem{yu2019meta}
\BIBentryALTinterwordspacing
T.~Yu, D.~Quillen, Z.~He, R.~Julian, K.~Hausman, C.~Finn, and S.~Levine, ``Meta-world: A benchmark and evaluation for multi-task and meta reinforcement learning,'' in \emph{Conference on Robot Learning (CoRL)}, 2019. [Online]. Available: \url{https://arxiv.org/abs/1910.10897}
\BIBentrySTDinterwordspacing

\bibitem{gupta2019relay}
A.~Gupta, V.~Kumar, C.~Lynch, S.~Levine, and K.~Hausman, ``Relay policy learning: Solving long-horizon tasks via imitation and reinforcement learning,'' \emph{arXiv preprint arXiv:1910.11956}, 2019.

\bibitem{cohen2022imitation}
\BIBentryALTinterwordspacing
S.~Cohen, B.~Amos, M.~P. Deisenroth, M.~Henaff, E.~Vinitsky, and D.~Yarats, ``Imitation learning from pixel observations for continuous control,'' 2022. [Online]. Available: \url{https://openreview.net/forum?id=JLbXkHkLCG6}
\BIBentrySTDinterwordspacing

\bibitem{kalashnikov2021mt}
D.~Kalashnikov, J.~Varley, Y.~Chebotar, B.~Swanson, R.~Jonschkowski, C.~Finn, S.~Levine, and K.~Hausman, ``Mt-opt: Continuous multi-task robotic reinforcement learning at scale,'' \emph{arXiv preprint arXiv:2104.08212}, 2021.

\bibitem{xie2022lifelong}
A.~Xie and C.~Finn, ``Lifelong robotic reinforcement learning by retaining experiences,'' in \emph{Conference on Lifelong Learning Agents}.\hskip 1em plus 0.5em minus 0.4em\relax PMLR, 2022, pp. 838--855.

\bibitem{kingma2014adam}
D.~P. Kingma and J.~Ba, ``Adam: A method for stochastic optimization,'' \emph{arXiv preprint arXiv:1412.6980}, 2014.

\bibitem{wang2020minilm}
W.~Wang, F.~Wei, L.~Dong, H.~Bao, N.~Yang, and M.~Zhou, ``Minilm: Deep self-attention distillation for task-agnostic compression of pre-trained transformers,'' 2020.

\bibitem{reimers-2019-sentence-bert}
\BIBentryALTinterwordspacing
N.~Reimers and I.~Gurevych, ``Sentence-bert: Sentence embeddings using siamese bert-networks,'' in \emph{Proceedings of the 2019 Conference on Empirical Methods in Natural Language Processing}.\hskip 1em plus 0.5em minus 0.4em\relax Association for Computational Linguistics, 11 2019. [Online]. Available: \url{https://arxiv.org/abs/1908.10084}
\BIBentrySTDinterwordspacing

\bibitem{wu2023mtm}
P.~Wu, A.~Majumdar, K.~Stone, Y.~Lin, I.~Mordatch, P.~Abbeel, and A.~Rajeswaran, ``Masked trajectory models for prediction, representation, and control,'' in \emph{International Conference on Machine Learning}, 2023.

\bibitem{caruana1997multitask}
R.~Caruana, ``Multitask learning,'' \emph{Machine learning}, vol.~28, pp. 41--75, 1997.

\bibitem{majumdar2023we}
A.~Majumdar, K.~Yadav, S.~Arnaud, Y.~J. Ma, C.~Chen, S.~Silwal, A.~Jain, V.-P. Berges, P.~Abbeel, J.~Malik, \emph{et~al.}, ``Where are we in the search for an artificial visual cortex for embodied intelligence?'' \emph{arXiv preprint arXiv:2303.18240}, 2023.

\bibitem{parisotto2015actor}
E.~Parisotto, J.~L. Ba, and R.~Salakhutdinov, ``Actor-mimic: Deep multitask and transfer reinforcement learning,'' \emph{arXiv preprint arXiv:1511.06342}, 2015.

\bibitem{xu2020knowledge}
Z.~Xu, K.~Wu, Z.~Che, J.~Tang, and J.~Ye, ``Knowledge transfer in multi-task deep reinforcement learning for continuous control,'' \emph{Advances in Neural Information Processing Systems}, vol.~33, pp. 15\,146--15\,155, 2020.

\bibitem{chen2018gradnorm}
Z.~Chen, V.~Badrinarayanan, C.-Y. Lee, and A.~Rabinovich, ``Gradnorm: Gradient normalization for adaptive loss balancing in deep multitask networks,'' in \emph{International conference on machine learning}.\hskip 1em plus 0.5em minus 0.4em\relax PMLR, 2018, pp. 794--803.

\bibitem{thrun1998lifelong}
S.~Thrun, ``Lifelong learning algorithms.'' \emph{Learning to learn}, vol.~8, pp. 181--209, 1998.

\bibitem{parisi2019continual}
G.~I. Parisi, R.~Kemker, J.~L. Part, C.~Kanan, and S.~Wermter, ``Continual lifelong learning with neural networks: A review,'' \emph{Neural networks}, vol. 113, pp. 54--71, 2019.

\bibitem{rebuffi2017icarl}
S.-A. Rebuffi, A.~Kolesnikov, G.~Sperl, and C.~H. Lampert, ``icarl: Incremental classifier and representation learning,'' in \emph{Proceedings of the IEEE conference on Computer Vision and Pattern Recognition}, 2017, pp. 2001--2010.

\bibitem{power2017neural}
J.~D. Power and B.~L. Schlaggar, ``Neural plasticity across the lifespan,'' \emph{Wiley Interdisciplinary Reviews: Developmental Biology}, vol.~6, no.~1, p. e216, 2017.

\bibitem{barnett2002and}
S.~M. Barnett and S.~J. Ceci, ``When and where do we apply what we learn?: A taxonomy for far transfer.'' \emph{Psychological bulletin}, vol. 128, no.~4, p. 612, 2002.

\bibitem{li2017learning}
Z.~Li and D.~Hoiem, ``Learning without forgetting,'' \emph{IEEE transactions on pattern analysis and machine intelligence}, vol.~40, no.~12, pp. 2935--2947, 2017.

\bibitem{mendez2018lifelong}
J.~Mendez, S.~Shivkumar, and E.~Eaton, ``Lifelong inverse reinforcement learning,'' \emph{Advances in neural information processing systems}, vol.~31, 2018.

\bibitem{chen2018neural}
R.~T. Chen, Y.~Rubanova, J.~Bettencourt, and D.~K. Duvenaud, ``Neural ordinary differential equations,'' \emph{Advances in neural information processing systems}, vol.~31, 2018.

\bibitem{auddy2023continual}
S.~Auddy, J.~Hollenstein, M.~Saveriano, A.~Rodr{\'\i}guez-S{\'a}nchez, and J.~Piater, ``Continual learning from demonstration of robotics skills,'' \emph{Robotics and Autonomous Systems}, p. 104427, 2023.

\bibitem{vitter1985random}
J.~S. Vitter, ``Random sampling with a reservoir,'' \emph{ACM Transactions on Mathematical Software (TOMS)}, vol.~11, no.~1, pp. 37--57, 1985.

\end{thebibliography}
